%% file: KneiflEtAl24Multi.tex
\documentclass[preprint,10pt]{elsarticle}

\usepackage[margin=1in]{geometry}
\pdfoutput=1 
\usepackage{amsmath}
\usepackage{amssymb}
\usepackage{subcaption}
\usepackage[capitalise]{cleveref}
\usepackage{tikz}
\usepackage{siunitx}
\usepackage{leftidx}
\usepackage{booktabs}
\usepackage{tabularx}
\usepackage{numprint}
\usepackage{makecell}
\usepackage{ifthen}
\usepackage{url}

\usetikzlibrary{external, positioning, calc, shapes, arrows}
\graphicspath{{fig/models/arm/}, 
    {fig/models/kart/}, 
    {fig/models/kart/simulations},
    {fig/results/kart/}, 
    {fig/results/kart/contribution/}}
\usepackage{pgfplots}
    \pgfplotsset{
        table/search path={fig/results/kart/data/},
    }
    \usepgfplotslibrary{groupplots,dateplot}
\input{mh_variables}

\input{fig/figure_config}

\newboolean{overleaf}
\setboolean{overleaf}{true}

\begin{document}
\makeatletter
\def\ps@pprintTitle{%
  \let\@oddhead\@empty
  \let\@evenhead\@empty
  \let\@oddfoot\@empty
  \let\@evenfoot\@oddfoot
}
\makeatother
\begin{frontmatter}


\title{Multi-Hierarchical Surrogate Learning for\\Structural Dynamical Crash Simulations\\Using Graph Convolutional Neural Networks}

\author[US]{Jonas Kneifl\corref{cor1}}
\cortext[cor1]{Corresponding author.}
\ead{jonas.kneifl@itm.uni-stuttgart.de}
\author[US]{J\"org Fehr}
\ead{joerg.fehr@itm.uni-stuttgart.de}
\author[UW1]{Steven L. Brunton}
\ead{sbrunton@uw.edu}
\author[UW2]{J. Nathan Kutz}
\ead{kutz@uw.edu}

\affiliation[US]{
    organization={Institute of Engineering and Computational Mechanics, University of Stuttgart},
    addressline={Pfaffenwaldring 9}, 
    city={Stuttgart},
    postcode={70569}, 
    state={Baden-Württemberg},
    country={Germany}}
\affiliation[UW1]{organization={Department of Mechanical Engineering, University of Washington},
            city={Seattle},
            postcode={98195}, 
            state={WA},
            country={USA}
            }
\affiliation[UW2]{organization={Department of Applied Mathematics and Electrical and Computer Engineering, University of Washington},
            city={Seattle},
            postcode={98195}, 
            state={WA},
            country={USA}
            }

\begin{abstract}
Crash simulations play an essential role in improving vehicle safety, design optimization, and injury risk estimation.
Unfortunately, numerical solutions of such problems using state-of-the-art high-fidelity models require significant computational effort. Conventional data-driven surrogate modeling approaches create low-dimensional embeddings for evolving the dynamics in order to circumvent this computational effort.  
Most approaches directly operate on high-resolution data obtained from numerical discretization, which is both costly and complicated for mapping the flow of information over large spatial distances. Furthermore, working with a fixed resolution prevents the adaptation of surrogate models to environments with variable computing capacities, different visualization resolutions, and different accuracy requirements.
We thus propose a multi-hierarchical framework for structurally creating a series of surrogate models for a kart frame, which is a good proxy for industrial-relevant crash simulations, at different levels of resolution.  For multiscale phenomena, macroscale features are captured on a coarse surrogate, whereas microscale effects are resolved by finer ones. The learned behavior of the individual surrogates is passed from coarse to finer levels through transfer learning. 
In detail, we perform a mesh simplification on the kart model to obtain multi-resolution representations of it. We then train a graph-convolutional neural network-based surrogate that learns parameter-dependent low-dimensional latent dynamics on the coarsest representation. Subsequently, another, similarly structured surrogate is trained on the residual of the first surrogate using a finer resolution. This step can be repeated multiple times. By doing so, we construct multiple surrogates for the same system with varying hardware requirements and increasing accuracy.
\end{abstract}

\end{frontmatter}


\section{Introduction}\label{sec1}

Improved passive and active safety systems in vehicles, such as airbags or optimized design and materials, significantly reduces fatalities in automotive accidents~\cite{NSC22}. Accordingly, safety considerations significantly shape the design of modern vehicles.
The use of of real cars or physical dummies alone is not practical for crashworthiness testing since they are prohibitively expensive, inflexible, and time-consuming to build and test. Thus, the safety enhancement of vehicles would be impractical without simulations making computer-aided engineering (CAE) a key pillar in crashworthiness development~\cite{Kramer2023}. Specifically, structural dynamical crash simulations are industry relevant and crucial for determining weaknesses in crash structure design and the optimization of material use. 

Crash simulations are used in a variety of applications including many-query evaluation, design optimization, or on-board computer evaluations. Many of these applications require high accuracy. However, classical numerical modeling methods are such that increased accuracy comes at the expense of increased computational effort. 
This means that modern high-fidelity simulation models of crash simulations and other technical or biological systems, while providing insight into complex dynamic processes and the ability to predict their behavior, have prohibitive computational costs for large parameter studies, for use on weak hardware, or for real-time applications. This leads to a critical need for surrogate models that keep their high-fidelity counterparts' expressiveness while being more cost-effective to simulate~\cite{Kneifl2021,Czech2022, Lesjak2023}. 

Surrogate modeling for crash simulations poses special challenges: 
Simulations often include multiple sophisticated parameter dependencies defined by diverse scenarios, contain complex material properties, and are subject to high nonlinearities and contact with plastic deformations. 
In contrast to other structural dynamical systems often considered in the literature, transient dynamics without stationary system behavior must be approximated instead of the post-transient dynamics on an attracting submanifold.
Moreover, they are usually modeled using finite element methods and consequently share the challenges with other structural dynamical systems including (i) the sheer dimensionality of such systems, (ii) the inaccessibility of commercial software code, and (iii) the computationally intensive data generation.

Data-based \emph{model order reduction} (\MOR) has emerged as a viable solution to the task of creating efficient yet accurate surrogate models, addressing two main problems~\cite{Noack2011book,Benner2015siamreview,Taira2017aiaa,Taira2020aiaaj,Brunton2020arfm,Brunton2022book}. One is the identification of expressive coordinates that are simultaneously low-dimensional to ensure computational efficiency, yet still adequate for describing the system. The other is the approximation of the (parameter-dependent) system dynamics in the identified reduced coordinates. 

The accuracy of classic solvers for crash simulations and other high-fidelity simulation models is limited by the resolution of the spatial discretization. Consequently, their dimensionality is driven by the spatial convergence, whereby the accuracy increases monotonically with the resolution. Fortunately, this means that the intrinsic dimension of the given problem is usually much smaller and the actual solution space lies on a low-dimensional manifold, which allows MOR methods to find suitable low-dimensional descriptions. 
Widely used data-driven methods to construct a low-dimensional embedding include linear methods such as the \emph{proper orthogonal decomposition} (\POD)~\cite{Volkwein2013} (also known as principal component analysis (\PCA)) and its nonlinear counterpart, the autoencoder (\AE), which produces nonlinear manifolds. 
Even though POD-based surrogate models continue to be widely used and are able to produce satisfying results for many problems~\cite{Hesthaven2018, Wang2019, Guo2019, Kneifl2021,Kneifl2022}, autoencoders~\cite{LeeCarlberg2020, Fresca2021, Kneifl2023} have been shown to outperform their linear counterpart for problems with slowly decaying Kolmogorov $n$-width~\cite{Peherstorfer2022}. 
Combinations of \POD and autoencoders~\cite{Fresca2022} are also possible. Of the many different autoencoder architectures, convolutional autoencoders in particular have stood out~\cite{GonzalezBalajewicz2018, LeeCarlberg2020,  Fresca2021, Maulik2021} as they exploit spatial information and can detect local patterns using filters. 
This also makes convolutional neural networks interesting for other applications in structural dynamics, see e.g.~\cite{Stoffel2020, Bamer2021}.
Meanwhile, \MOR methods such as \POD have also been used to improve convolutional neural networks, for example to reduce the number of layers~\cite{Meneghetti2023}.

Unfortunately, conventional \emph{convolutional neural networks} (\CNNs) face the significant limitation of only being applicable to regular grid-like data (e.g. images). Irregular data, on the contrary, as it is present in complex three-dimensional discretized crash simulations requires new techniques, thus leading to an increased interest in geometrical deep learning~\cite{Bronstein2021}. While some approaches map the irregular domain to a regular one~\cite{GaoSunWang2021} and apply convolutions there, others apply convolutional-like operations on dynamically constructed graphs of point-clouds~\cite{WangEtAl2019}, and still others apply generalizations of CNN architectures to non-Euclidean domains~\cite{MontiEtAl2017}. \emph{Graph convolutional neural networks}~(\GCNNs)~\cite{Zhou2020} can be directly applied to irregular data, by transferring the principle of convolutions to geometrical problems. They can extract information and relations about features of nodes from their spatial connections. An early version of \GCNNs can be found in \cite{DefferrardBressonVandergheynst2016} and an adaption of it in~\cite{KipfWelling2016}.

GCNNs are found in the context of~\MOR in \cite{GruberEtAl2022}, where a graph convolutional autoencoder using gcn2 convolutions~\cite{ChenEtAl2020} is compared to a classic fully connected \AE for the creation of reduced order models. Furthermore, in~\cite{Pichi2024} a spatial graph convolutional autoencoder is used to derive reduced order models and~\cite{Franco2023} utilizes~\GCNNs for the approximation of time-dependent PDEs under geometric variability.
In contrast to classic convolutions, graph convolutions themselves cannot automatically reduce the dimensionality of the data. To decrease the number of nodes that are processed in the layers, several pooling operations for irregular data are developed~\cite{GrattarolaEtAl2021}.
A general overview of \GCNNs can be found in~\cite{WuEtAl2021}, with a literature review focused on MOR in~\cite{Pichi2024}. 
\enlargethispage{\baselineskip}

In addition to graph convolutional neural networks, many other exciting applications in geometrical deep learning have emerged like graph networks~\cite{Battaglia2018}. Graph network-based simulation models have been used to model a physically informed simulation model in which the graph network outputs state-time derivatives that are then used in ODE integrators for future time predictions~\cite{Pfaff2020, SanchezEtAl2020}.  Moreover, symbolic representations of a learned model are discovered by applying symbolic regression to components of its message passing function~\cite{Cranmer2019,CranmerEtAl2020}. Graph networks are also used in generative tasks where graphs are built sequentially based on learned distributions~\cite{Li2018}.

Most of the \MOR approaches directly operate on the given high-dimensional discretization, i.e. the resulting surrogates always try to approximate a system for a fixed high resolution. However, as already mentioned this resolution usually is neither driven by the underlying problem nor the user's intended application. In many cases, a coarser or adaptive resolution has advantages if the accuracy does not suffer as a result. For example, in visualization of complex three-dimensional systems significant computational power can be saved using coarser resolutions. Moreover, static resolutions cannot react to changes in computational environments, like dynamically changing memory restrictions, or to changes in the desired approximation quality. Consequently, the question arises as to why we should limit data-based surrogate models to these fixed original resolutions when they have the advantage of not being limited to spatial convergence? They can operate on a coarse subselection of the fine mesh with an accuracy only limited by their expressiveness and the high-fidelity data. One discretization-free approximation scheme for parametrized PDEs can be found in \cite{ChenEtAl2022} and other mesh-free approaches are given in~\cite{rodriguez2022projection} and \cite{li2020fourier,li2020neural}.

In this work, we develop an approach that is fundamentally mesh-free, i.e., it is not restricted to the underlying high-resolution discretization. Instead, we take advantage of the fact that the surrogates do not require fine spatial resolution by first excluding large parts of the model during model creation while taking into account the recent advantages of \GCNNs. 
Therefore, we transfer and adjust ideas from multiresolution autoencoders~\cite{Liu2023} to make them applicable to irregular data.
In particular, we present a graph-convolutional hierarchical multiscale approximation scheme for a given system in which the global context is captured in coarse representations. By doing so, we can speed up the learning process, create multiple models with individual hardware requirements, and resolve multiscale issues that often arise in spatio-temporal dynamics of complex systems. 

\begin{figure}
    \centering
    \ifthenelse{\boolean{overleaf}}
        {
            \includegraphics[width=.95\textwidth]{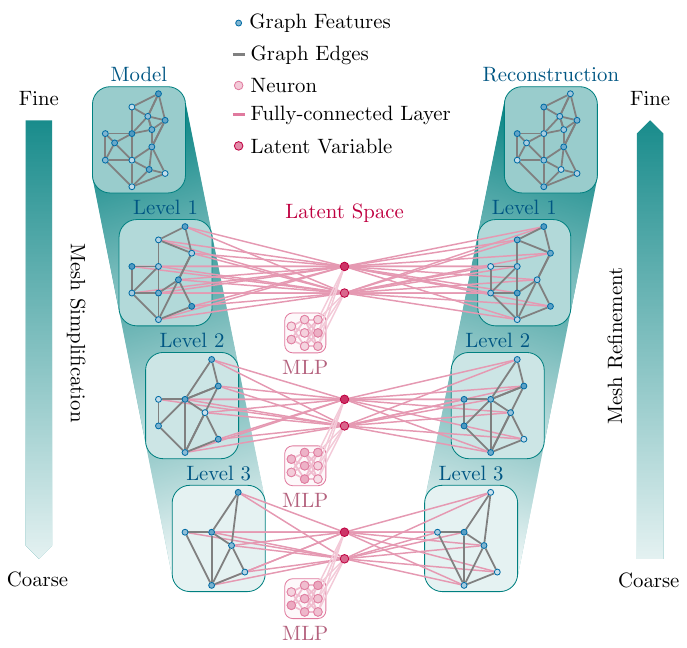}
        }
        {
            \input{fig/methods/multi_hierarchic/multi_hierarchic_surrogate_modeling.tex}
        }
    \caption{Multi-hierarchical Surrogate modeling approach. Instead of learning on the full system discretization, we learn on a coarse representation of it. We then progressively refine the coarse representation by learning on the residual error. First, different levels of discretization are generated for the original model. On the coarsest level, a surrogate model is trained. If the error is not within the tolerance, the learned model is upsampled to the next level, and an additional surrogate is learned to capture the inaccuracies of the first one. The process is repeated until the error is within the tolerance or no more discretizations are left.}
    \label{fig:workflow}
\end{figure}

Our fundamental idea is (i) to represent the high-fidelity model in a graph-like structure, (ii) apply mesh simplification to derive coarse representations, (iii) fit a surrogate model on the coarsest representation, (iv) refine the model, and (v) fit another surrogate on the next finer level leveraging transfer learning. The steps (iv) and (v) can be repeated until a performance threshold is met or no more coarse representations are available. An abstracted visual impression of this workflow is given in \cref{fig:workflow}, where it may be seen that our approach resembles U-NETs~\cite{ronneberger2015u} in its structure. The individual surrogate models themselves are composed out of a graph convolutional autoencoder which constructs low-dimensional coordinates, and out of a multi-layer perceptron (\MLP) that approximates those coordinates based on the time and given parameters. 

Due to the hierarchical structure, global dynamics can be captured on the coarsest surrogate, whereas finer details are captured in the refined versions. The framework is naturally suited for multiscale problems, where macro- and microscale dynamics occur at the same time. Modeling such systems poses a particular challenge and is therefore often approached in a special way. For example, in multigrid methods~\cite{multigrid87, Trottenberg2000}, where a coarse-grained model is gradually refined in areas of high inaccuracy in order to achieve the required accuracy. Such methods also have been unified with convolutional neural networks \cite{HeXu2019}.

Including hierarchical structure in \GCNNs is a natural way of proceeding. In Graph U-Nets~\cite{GaoEtAl2021}, pooling layers are used to form smaller graphs using a trainable projection vector. Moreover, a multiscale MeshGraphNet that operates on different resolutions is introduced in~\cite{FortunatoEtAl2022}. A coarse resolution is used to propagate information further and overcome the issue of slow message propagation occurring in fine resolutions. Mesh reduction is applied in~\cite{HanEtAl2022} to encode information of a fine graph in a coarse subset of the nodes. By doing so they can evolve the latent dynamics efficiently in time using an attention model.
Note that the hierarchical approaches are also incorporated in other architectures like variational autoencoders~\cite{Lee2023} and are not only used in the spatial domain but also for the evolution of dynamics in time, as explained in~\cite{Liu2022}.

However, all those approaches use hierarchical representations only to foster the learning process for an approximation of the fine discretized high-fidelity solution. In contrast, we use coarse representations that are physically and visually interpretable and consequently directly useful. Moreover, we never perform costly training of a surrogate on the fine data and can reduce the latent representation to a vector of the intrinsic dimension instead of a small but comparably larger graph. All these steps are taking us towards digital twins~\cite{niederer2021scaling} for crash simulations.
Additionally, we build the surrogates one after the other enabling the learning process to be stopped if desired. In doing so, we still take advantage of already learned behavior by applying transfer learning from coarser to refined surrogates. 
The highlights of our work can be summarized as follows:
\begin{enumerate}
    \item Create accurate yet efficient data-driven surrogate for a complex nonlinear crash simulation model in a structured and efficient way.
    \item Develop a multiresolution surrogate architecture for complex discretized 3D structural dynamical problems.
    \item Transfer learning is leveraged to progressively refine the surrogate model in a hierarchical fashion.
    \item Create adaptive models for different needs (resolution, memory etc.) in visually interpretable domains.
\end{enumerate}

The paper is structured as follows: The considered problem scenario of a crash simulation model is introduced in \cref{sec: problem}, followed by the proposed multi-hierarchical surrogate modeling approach in \cref{sec: theory} along with the required theory. Its application as well as the results and discussion are presented in \cref{sec: results}. The paper ends with a conclusion in \cref{sec: conclusion}.

\section{A Crashworthiness Simulation Example}\label{sec: problem}

Crash simulations play a fundamental role in safety evaluation, safety system design optimization, or on-board evaluation. These applications require multi-query evaluations, real-time capability or computational efficiency. 
In this work, we consider a simplified crash model that takes into account industrially relevant challenges -- as explained in the introduction section -- but is also suitable for research purposes. It offers all the interesting aspects of scenario variations with multiple complex parameter dependencies, nonlinear material and contact behavior with plastic deformations, while still being computationally tractable and easy to comprehend.

\subsection{A Racing Kart Frontal Crash Simulation}
\begin{figure}
    \centering
    \ifthenelse{\boolean{overleaf}}
    {
        \includegraphics[]{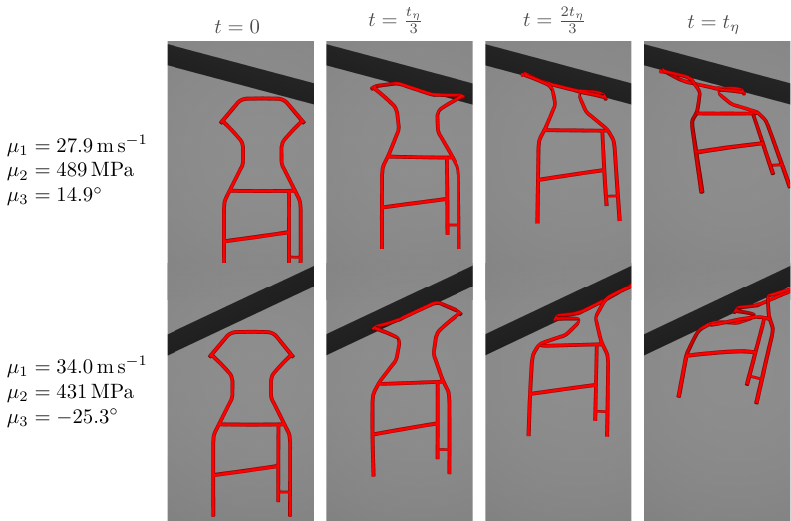}
    }
    { 
        \input{fig/models/kart/simulations/kart_crash_simulations.tex}
    }
	\caption{Two example simulations of the the kart frame~\cite{Shiiba2012} that serves as example of a structural mechanical crash system. From each simulation four snapshots at different points in time are used for the visualization.}
	\label{fig: kart crash}
\end{figure}

The high-fidelity model considered in the following experiments represents the frame of a racing kart, which is responsible for the essential dynamic behavior of a kart~\cite{Shiiba2012}. 
The remaining bodies of a kart like its wheels, vehicle shell, or engine are not considered in order to render a more tractable model.
Slight variations of this model have already been used in~\cite{Fehr2016, Kneifl2021}. 
The frame itself is a finite element model implemented in the commercial software tool LS-Dyna and consists of~$\nNodes=9314$ nodes with~$\nChannels^{(0)}=3$ translational degrees of freedom each.
The considered simulation scenario describes a crash of the kart against a rigid wall under varying impact speed~$\param_1\in[5, 35]$\qty{}{\metre\per\s}, impact angle~$\param_2\in[-45, 45]$\qty{}{\degree}, and yield stress~$\param_3\in[168,  758]$\qty{}{\mega\pascal}. 
The impact angle describes the angle between the normal of the wall and the orientation of the kart whereas the yield stress impacts the effective plastic stress-strain curve of the kart's material. The curve corresponds to one typical for steel and is offset by the individual yield stress of each simulation.
Each crash simulation covers a simulation time of~\qty{0.003}{\second} with a sampling time of~\qty{0.3}{\milli\second} resulting in 101 samples per simulation. Two example simulations are showcased in \cref{fig: kart crash}.
The major goal of our paper is to derive a surrogate model that can reproduce the high-fidelity simulation results of the kart in multiple resolutions with high accuracy and low computational times. Conventional approaches reach their limits in doing so due to the model's complexity.

\begin{figure}[t!]
    \centering   
    \ifthenelse{\boolean{overleaf}}
    {
        \includegraphics[]{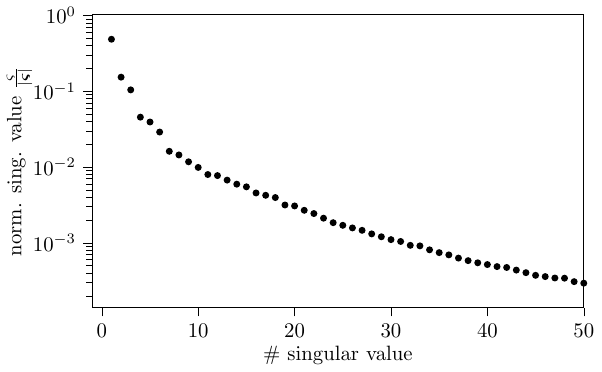}
    }
    {
        \input{fig/results/kart/singular_values/sing_values.tex}
    } 
    \hfill
    \caption{The first 50 normalized singular values of the training data as indicator of the Kolmogorov n-width.}
    \label{fig: singular values}
\end{figure}

\paragraph*{\textit{Model complexity}}
To showcase the complexity of the presented crash simulation model in the context of MOR, we consider the course of the normalized singular values of the high-fidelity simulation results in \Cref{fig: singular values}. The data matrix is composed of snapshots of the system states~$\states\in \stateSpace \subseteq \Rdim^{\stateDim}$.
The magnitude of each singular value reflects the importance of the corresponding reduced basis vector for describing the data. If a few singular values are dominant, then the data can be described well with a linear combination of only a few reduced basis vectors.  If not, a non-negligible error is introduced or more basis vectors must be used. Accordingly, the singular values can serve as an indicator for the Kolmogorov $n$-width~(\cite{LeeCarlberg2020}, \cite{Unger2019})
\begin{align*}
	d_n(\Flow(\paramSpace))
	:=\underset{\stateSpace_{n} \subseteq \stateSpace}{\inf} \ 
		\underset{\states\in\solutionManifold_{\paramSpace}}{\sup} \ 
			\underset{\stateApprox_n\in\stateSpace_{n}}{\inf}
			\| \states-\stateApprox_n \|,
\end{align*}
which quantifies the optimal linear trial subspace by describing the largest distance between any point in the solution manifold~$\solutionManifold_{\paramSpace}$ for all parameters and all $n$-dimensional subspaces~$\stateSpace_{n}\subseteq \stateSpace$.
For the considered problem the intrinsic dimension of the solution space is at most equal to the number of parameters plus one for the time resulting in~$\redStateDim=4$.
However, since not only the first four but also the subsequent singular values make significant contributions, it can be assumed that linear reduction methods as \PCA lead to appreciable errors. Hence, we introduce a more sophisticated \MOR approach to generate high-quality approximations in the following section.

\section{Multi-hierarchical Surrogate Modeling Approach}\label{sec: theory}
A detailed explanation of the general problem setup, the theory required to follow the explanations and most importantly, the multi-hierarchical surrogate modeling approach itself is given in this section. Although we especially apply the proposed methods to crash simulations, the explanations follow a general manner, so the interested readers can transfer them to their individual problem classes more easily.

\subsection{Problem Setup}
Consider a nonlinear dynamical system
\begin{align}
    \ddt\states(\timee, \params)=\rhs(\states(\timee, \params), \timee, \params)
    \label{eq: dynamical system}
\end{align}
which is determined by the time~$\timee\in\timeSpace\subseteq \Rdim^{+}$, the system state~$\states\in \stateSpace \subseteq \Rdim^{\stateDim}$, and (simulation) parameter~$\params\in \paramSpace \subseteq \Rdim^{\paramDim}$. Usually, time-stepping schemes are used to approximate the discrete-time \emph{flow} map 
\begin{align}
    \states(\timee, \params)=\Flow(\timee,\params,\states_{0}),
    \label{eq:flow map}
\end{align}
i.e. the solution of system~\cref{eq: dynamical system} at~$\nTime$ discrete time points~$\timee\in\{\timee_0,...,\timee_{\nTime-1}\}$. The flow map~$\Flow:\timeSpace \times \paramSpace \times \stateSpace \to \stateSpace$ describes the mapping from the initial condition~$\states_0\in\Rdim^{\stateDim}$ and parameter~$\params$ to the solution at a given time~$t\geq t_0$. 
In the course of this paper, our major goal is to find a surrogate model~$\surrogate$ that approximates the solution~\cref{eq:flow map} of~\cref{eq: dynamical system}, i.e.
\begin{align*}
    \stateApprox(\timee, \params)=\surrogate(\timee,\params,\states_{0})\approx\Flow(\timee,\params,\states_{0})
\end{align*}
while significantly reducing computational requirements. This can often be expressed in terms of the computational time so that the computational time of the surrogate~$\compTime_{\surrogate}$ is much faster than that of the original system~$\compTime_{\Flow}$: $\compTime_{\surrogate} \ll \compTime_{\Flow}$.

If, however, not all states are of interest but only a subselection, which is the case when coarsening the original discretization of the system, the surrogate can operate on a downsampled state~$\states_{\down}(\timee, \params)=
\downsampling\states\in\stateSpace_{\down} \subset \stateSpace$,~$\stateSpace_{\down} \subset \Rdim^{\nNodesDown}$, where~$\downsampling \in \binare^{\nNodesDown \times \stateDim}: \stateSpace \to \stateSpace_\down$ is a binary selection matrix with~$\nNodesDown < \nNodes$. Its entries are~$\downsampling(\nodeIteratorA,\nodeIteratorB)=1$ when the~$\nodeIteratorB$-th state is kept or $\downsampling(\nodeIteratorA,\nodeIteratorB)=0, \ \forall \nodeIteratorA\in\{1,\dots,\nNodesDown\}$ when the ~$\nodeIteratorB$-th state is discarded.
Consequently, the surrogate~$\surrogate_\down$ only needs to approximate the selected states of the system's solution
\begin{align*}
    \stateApprox_\down(\timee, \params)=\surrogate_\down(\timee,\params,\downsampling\states_{0})\approx\downsampling\Flow(\timee,\params,\states_{0})=\states_\down(\timee, \params).
\end{align*}

This can on the one hand ease the surrogate modeling process, but is on the other hand especially useful in cases when the fine discretization of the model does not result  from user requirements. In order to obtain a suitable downsampling operation, methods from the field of computer vision are useful.

\subsection{Down- and Upsampling}\label{sec: down and upsampling}
For the sampling operations, we rely on surface simplification using quadric error metrics~\cite{Garland1997}, which is a method that produces coarse representations of a given mesh maintaining its shape, i.e., its geometrical characteristics. The method does not necessarily preserve the topology of the mesh as topological holes can be closed and unconnected regions can be joined. Classic \FE mesh simplification approaches focus on maintaining the topology, but as we are interested in coarse representations of the original system that have a similar visualization the mentioned method is preferable.
The same sampling approach is used as a pooling operation in the context of graph convolutional autoencoders in CoMA~\cite{Ranjan2018}.

Specifically, we assume that the considered model can be interpreted as an undirected graph~$\graph=(\nodes, \edges, \adjacency)$ with a set of vertices (nodes)~$\nodes\in\Rdim^{\nNodes \times 3}$ and edges~$\edges\in\Rdim^{\nEdges \times 2}$ describing the node connectivity defined by the adjacency matrix~$\adjacency\in\binare^{\nNodes \times \nNodes}$. Note that the adjacency can also be weighted.
The high-fidelity model used in this elaboration is a finite element~(\FE) model so that the representation as a graph corresponds to the model formulation, as \FE models are composed out of elements that contain nodes and define neighborhoods through their edges.

\paragraph*{\textit{Downsampling Operation}}
The downsampling operation of the nodes is defined by 
\begin{align*}
    \nodesDown=\downsampling\nodes
\end{align*}
with the downsampling matrix defined as previously but with dimensions aligning to the number of nodes~$\downsampling \in \binare^{\nNodesDown \times \nNodes}:\nodes \to \nodesDown$ with~$\nNodesDown < \nNodes$ and~$\nodesDown \subset \nodes$.
The selection of the nodes to keep follows~\cite{Garland1997} using iterative vertex pair contraction. In general, for a given pair of nodes~$(\node_\nodeIteratorA, \node_\nodeIteratorB)$, a vertex pair contraction $(\node_\nodeIteratorA, \node_\nodeIteratorB) \to \node$ moves node~$\node_\nodeIteratorB$ to a new position $\node$, connects incident edges to $\node_\nodeIteratorB$, deletes the second node~$\node_\nodeIteratorA$, and removes all degenerate edges and faces. As we only consider selected nodes instead of adjusted nodes, the position of the kept node is not changed, and the contraction results in~$(\node_\nodeIteratorA, \node_\nodeIteratorB) \to \node_\nodeIteratorB$. To, introduce a measure which determines which nodes are kept, 
each node~$\node=[\nodeComponent_x, \nodeComponent_y, \nodeComponent_z, 1]^\transpose$ is associated with an quadratic error 
\begin{align}
    \Delta(\node)=\node^\transpose(\quadric)\node
    \label{eq: quadric error}
\end{align}
which is defined w.r.t~$\quadric\in\Rdim^{4 \times 4}$ describing the distance of a given point to the set of planes on which intersections the node is placed.
The procedure can be summarized as follows:
\begin{enumerate}
    \item Select valid vertex pairs (either neighbours or close distant nodes)
    \item Select best~$\node$ out of~$\{\node_\nodeIteratorA, \node_\nodeIteratorB\}$ for each valid pair based on cost~$\node^\transpose(\quadric_\nodeIteratorA+\quadric_\nodeIteratorB)\node$
    \item Iteratively remove node from pair~$(\node_\nodeIteratorA, \node_\nodeIteratorB)$ of least cost, update costs
\end{enumerate}

\begin{figure}
    \centering
    \ifthenelse{\boolean{overleaf}}
    {
        \includegraphics[]{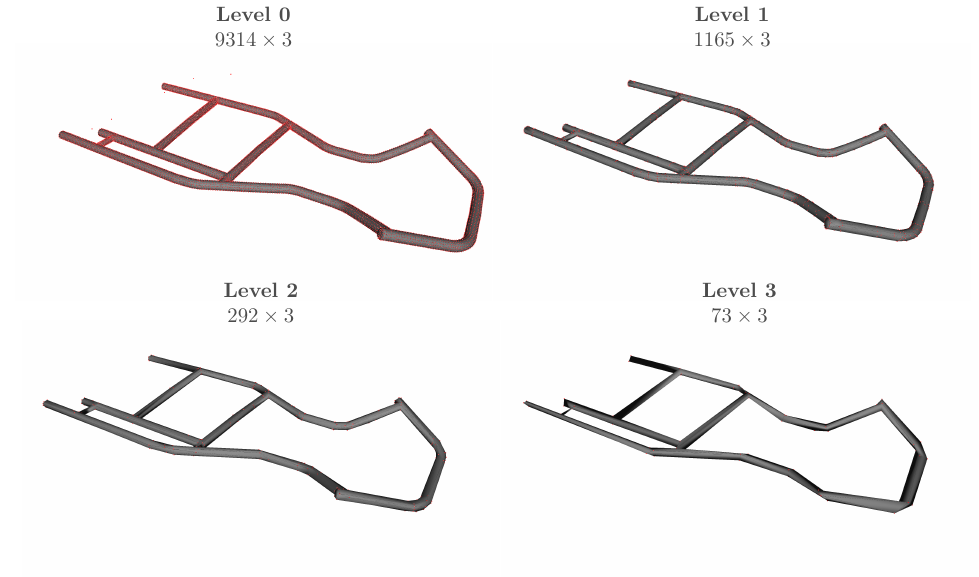}
    }
    {
        \input{fig/models/kart/downsampling_kart.tex}
    }
	\caption{Differently resolved representations of the kart frame. The red dots represent the nodes.}
	\label{fig: levels}
\end{figure}

\paragraph*{{Upsampling Operation}}
It is of interest to recover the original representation of the model from every coarse representation given. Unfortunately, a lossless reconstruction of the original mesh based on the simplified one is in general not possible. Consequently, we seek an upsampling matrix $\upsampling \in \Rdim^{\nNodes \times \nNodesDown}$ with
\begin{align}
    \nodesUp=\upsampling\nodesDown
    \label{eq: upsampling}
\end{align}
that approximates the original mesh. In this work, we follow the procedure of~\cite{Ranjan2018} and generate the upsampling matrix during the downsampling matrix creation process. 
A node~$\node_\nodeIteratorB$ that is kept in the downsampling process will lead to an entry in the upsampling matrix that follows~$\upsampling(\nodeIteratorA,\nodeIteratorB)=1$. A discarded node~$\node_\nodeIteratorA$, on the contrary, is mapped onto the down-sampled mesh using barycentric coordinates projecting it into the closest triangle $(i,j,k)$ in the down-sampled mesh 
\begin{align*}
    \tilde{\node}_\nodeIteratorA=\weight_i \node_i+\weight_j \node_j+\weight_k \node_k
\end{align*}
with $\node_i,\node_j,\node_k \in \nodesDown$ and $\weight_i+\weight_j+\weight_k=1$. 
The upsampling matrix is then updated with the corresponding weighting factors so that $\upsampling(\nodeIteratorA,i)=\weight_i, \ \upsampling(\nodeIteratorA,j)=\weight_j, \ \upsampling(\nodeIteratorA,k)=\weight_k$.
Visual examples of the coarsened \FE model is given in~\cref{fig: levels}.

\subsection{Surrogate Modeling}\label{subsubsec3}
Having the coarse representations of a model present, enables the surrogate modeling process at the different levels. In this work, we rely on graph convolutional neural networks~(\GCNNs) to create a low-dimensional latent representation of the system state. Note that any other data-driven dimensionality reduction scheme, linear as well as nonlinear, can replace the~\GCNNs. Nevertheless, \GCNNs profit from the proposed framework since the learning process is eased and accelerated, and they provide the best approximation quality among the tested methods. 

\subsubsection{Graph Convolutional Neural Networks}
Graph convolution neural networks generalize convolutional neural networks to irregular discretized domains as it is present in \FE models.
To understand the underlying principle of graph convolutions, recall a graph $\graph=\{\nodes,\edges,\adjacency\}$ as described in \cref{sec: down and upsampling}. A graph signal~$\mlInputs\in\Rdim^{\nNodes}$ is a feature vector of all~$\nNodes$ nodes in the graph. A beneficial technique to calculate a convolution between a filter~$\filter\in\Rdim^{\stateDim}$ and signal~$\mlInputs$ is that a convolution is just a multiplication in Fourier space. 
To receive a Fourier transform~$\fourier{\mlInputs}$ of the signal, a Fourier basis can be obtained from an eigenvalue factorization of the normalized Laplacian of a graph. The Laplacian is defined as~$\laplacian^*=\degreeMatrix-\adjacency$ with the adjacency matrix~$\adjacency=\adjacency(\graph)$ and the diagonal matrix of node degrees~$\degreeMatrix$ with entries~$\degreeMatrix_{i,i}=\sum_j \adjacency{i,j}$. The normalized version of the Laplacian~$\laplacian=\eye_n-\degreeMatrix^{-\frac{1}{2}}\adjacency\degreeMatrix^{\frac{1}{2}}$ is real symmetric positive semidefinite. Hence, the factorization $\laplacian=\eigenvectors \eigenvalueMatrix \eigenvectors^\transpose$ exists and the matrix~$\eigenvectors=\begin{bmatrix} \eigenvector_1 & \eigenvector_2 & \dots & \eigenvector_n \end{bmatrix}$ represents eigenvectors of the Laplacian ordered by their corresponding eigenvalues, which are stored in the diagonal matrix~$\eigenvalueMatrix$. 
The eigenvectors~$\eigenvector$ are known as Fourier modes of~$\graph$.

A Fourier transform of a signal~$\states$ is then given by~$\fourier{\states}=\Fourier(\states)=\eigenvectors^\transpose\states$ and the inverse Fourier transform by~$\states=\Fourier(\fourier{\mlInputs})^{-1}=\eigenvectors\fourier{\states}$.
Given those transformations, the graph convolution between the signal~$\mlInputs$ and a filter~$\filter$ results in
\begin{align}
    \mlInputs\ast\filter=\Fourier^{-1}(\Fourier(\mlInputs)\odot\Fourier(\filter))=\eigenvectors(\eigenvectors^\transpose\mlInputs\odot\eigenvectors^\transpose\filter),
    \label{eq: graph convolution}
\end{align}
where~$\odot$ represents the Hadamard/elementwise product.\\
Denoting the filter as~$\filterMatrix=\diag(\eigenvectors^\transpose\filter)$ and using the conversion~$\bm{a}\odot\bm{b}=\diag(\bm{b})\bm{a}$, \cref{eq: graph convolution} simplifies to
\begin{align}
    \mlInputs\ast\filterMatrix=\eigenvectors\filterMatrix\eigenvectors^\transpose\mlInputs,
    \label{eq: spectral graph convolutions}
\end{align}
which is a formulation all spectral-based \GCNNs follow. 
The idea in spectral convolutional neural networks is that the filters~$\filterMatrix=\Weights_{\channelIterator_i,\channelIterator_j}^{(\layerIterator)}=\diag(\weights_{\channelIterator_i,\channelIterator_j}^{(\layerIterator)})$ are the learnable weights~$\Weights_{\channelIterator_i,\channelIterator_j}^{(\layerIterator)}$ in convolutional layers
\begin{align}
    \mlInputMatrix_{:,\channelIterator_j}^{(\layerIterator+1)}=\activationFunction^{(\layerIterator)}(
        \sum_{\channelIterator_i=1}^{\nChannels^{(\layerIterator)}}
            \eigenvectors\Weights_{\channelIterator_i,\channelIterator_j}^{(\layerIterator)}\eigenvectors^\transpose \mlInputMatrix^{(\layerIterator)}_{:,\channelIterator}
    ).
    \label{eq: graph convolution layer}
\end{align}
Here, $\layerIterator$ denotes the layer index, $\channelIterator_i$ and $\channelIterator_j$ are the channel indices, $\nChannels$ is the number of channels in the $\layerIterator$-th layer, $\activationFunction$ is the activation function, $\Weights_{\channelIterator_i,\channelIterator_j}^{(\layerIterator)}$ is a diagonal matrix with learnable weights of the $\layerIterator$-th layer, and $\mlInputMatrix^{(\layerIterator)}_{:,\channelIterator}$ is the $\channelIterator$-th channel of $\mlInputMatrix^{(\layerIterator)}\in\Rdim^{\nNodes\times \nChannels^{(\layerIterator)}}$ where~$\mlInputMatrix^{(0)}=\snaps\in\Rdim^{\nNodes\times \nChannels^{(0)}}$ represents the original signal of the graph, in our case the~$\nNodes$ nodes of the \FE model with the coordinates stored in three channels, i.e. $\nChannels^{(0)}=3$. The filter formulation of \cref{eq: graph convolution layer} is not localized in space and requires a high learning complexity. 
Hence, the use of polynomial filters~$\filterMatrix=\sum_{\chebIterator}^{\chebOrder}\weight_\chebIterator\eigenvalueMatrix^\chebIterator$ is considered in~\cite{DefferrardBressonVandergheynst2016}. 
As such filters still require costly matrix multiplications with the non sparse Fourier basis~$\eigenvectors$ they propose to use polynomials that can be recursively calculated from the Laplacian~$\laplacian$ resulting in ChebNet~\cite{DefferrardBressonVandergheynst2016}.

\paragraph*{\textit{Chebyshev Spectral Convolutional Neural Networks}}
In Chebyshev spectral convolutional neural networks~\cite{DefferrardBressonVandergheynst2016}, the filter~$\filterMatrix$ is approximated by Chebyshev polynomials~$\chebPoly$ of order~$\chebOrder$. By doing so, the costly multiplications with the non sparse Fourier basis are replaced by~$\chebOrder$ multiplications with the sparse Laplacian.  
In detail, the filter is represented by the Chebyshev polynomials~$\filterMatrix(\scale{\eigenvalueMatrix})=
    \sum_{\chebIterator=0}^{\chebOrder} 
        \weights_\chebIterator \chebPoly_\chebIterator(\scale{\eigenvalueMatrix})$ 
of the eigenvalue matrix~$\scale{\eigenvalueMatrix}$ of the scaled Laplacian~$\scale{\laplacian}=2\laplacian / \eigenvalue_{\max} - \eye_n$.
Here, $\weights_\chebIterator$ are learnable polynomial coefficients, and the scaling ensures that all eigenvalues are within~$[-1,1]$.

Substituting, this filter in \cref{eq: spectral graph convolutions} and 
exploiting the transformation $\filterMatrix(\scale{\laplacian})=\eigenvectors\filterMatrix(\scale{\eigenvalueMatrix})\eigenvectors^\transpose$,
results in a graph convolution
\begin{align}
    \mlInputs\ast\filterMatrix& =
    \eigenvectors\filterMatrix(\scale{\eigenvalueMatrix})\eigenvectors^\transpose\mlInputs=
        \eigenvectors \left(
            \sum_{\chebIterator=0}^{\chebOrder} 
                \weights_\chebIterator \chebPoly_\chebIterator(\scale{\eigenvalueMatrix}) \right)
        \eigenvectors^\transpose\mlInputs=
        \sum_{\chebIterator=0}^{\chebOrder}\weights_{\chebIterator} \chebPoly_\chebIterator(\scale{\laplacian})\mlInputs
    \label{eq: cheb filter}
\end{align}
that gets rid of multiplication with~$\eigenvectors$. Chebyshev polynomials themselves can be recursively calculated as~$\chebPoly_\chebIterator(\bm{a})=2\bm{a}\chebPoly_{\chebIterator-1}(\bm{a})-\chebPoly_{\chebIterator-2}(\bm{a})$ with~$\chebPoly_{1}=\bm{a}$ and~$\chebPoly_{0}=\bm{0}$. 

In~\cite{KipfWelling2016}, the Graph Convolutional Network (GCN) is introduced which represents a first order approximation of ChebNet. Often GCNs face overfitting and oversmoothing, which was mitigated in GCN2~\cite{ChenEtAl2020}, where skip connections are used to propagate information over multiple layers. This approach is used in the context of \MOR in~\cite{GruberEtAl2022}. Nevertheless, ChebNet yields better results for our example and is consequently used in the following.

\subsubsection{Network Architecture: A Graph Convolutional Autoencoder with a Multilayer Perceptron}\label{sec:surrogate}
\begin{figure}
    \centering
    \ifthenelse{\boolean{overleaf}}
    {
        \includegraphics[]{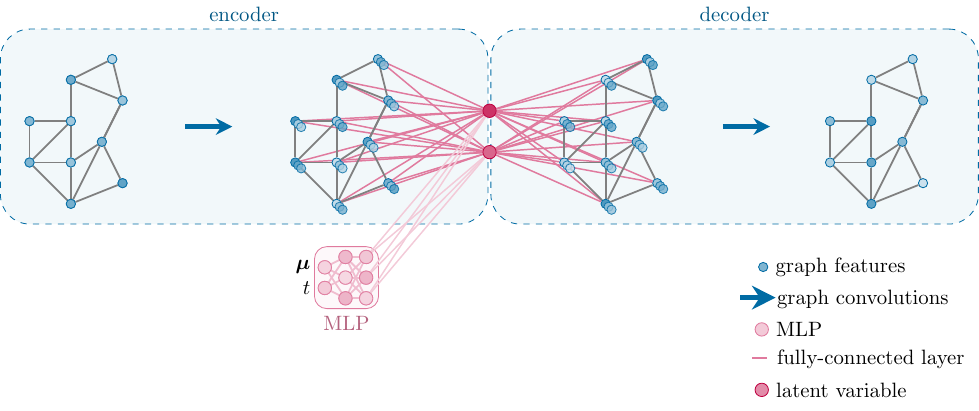}
    }
    {
        \input{fig/methods/NetworkArchitecture2/network_architecture.tex}
    }
    \caption{Graph convolutional autoencoder architecture with graph convolutional layers which operate on a mesh without pooling changing the number of features per node, fully-connected layers in the middle to reduce the dimensionality and an additional multilayer perceptron to capture the reduced dynamics.}
    \label{fig: network architecture}
\end{figure}

The network architecture that we use to create a surrogate model on the lowest level is shown in Figure~\ref{fig: network architecture}. It consists of (i) a (graph convolutional) autoencoder which is used to learn a low-dimensional embedding~$\redStateSpace\subseteq\Rdim^{\redStateDim}$ for the high-dimensional state space~$\stateSpace$ and (ii) a multilayer perceptron (\MLP) to capture the parameter- and time-dependencies in the identified low-dimensional latent manifold. 
The autoencoder consists of an encoder~$\reduction: \stateSpace \to \redStateSpace $ with learnable weights~$\Weights_{\text{enc}}$ mapping the high-dimensional state to a low-dimensional latent representation, i.e.~$\redState = \reduction(\states, \Weights_{\text{dec}})$, and a decoder~$\reconstruction: \redStateSpace \to \stateSpace$ with learnable weights~$\Weights_{\text{dec}}$ reconstructing the high-dimensional state from the low-dimensional latent representation, i.e.~$\stateRec = \reconstruction(\redState, \Weights_{\text{dec}})$. 
In case of graph convolutional layers, $\Weights_{\text{enc}}$ and~$\Weights_{\text{dec}}$ contain the trainable filters. 
The multilayer perceptron~$\regression: \paramSpace \times \timeSpace \times \redStateSpace \to \redStateSpace$ maps the parameters, time and the encoded initial condition~$\redState_0$ to the corresponding latent state~$\redStateApprox = \regression(\params, \timee, \redState_0, \Weights_{\text{mlp}})$ with trainable weights~$\Weights_{\text{mlp}}$. As we only consider simulations starting from the same initial condition in our example, it is neglected in the following.

The complete autoencoder reconstructs a state following
\begin{align}
    \stateRec=\reconstruction \circ \reduction(\states)
    \label{eq: autoencoder}
\end{align}
and the surrogate model that captures the (parametric) system dynamics is a function composition of the \MLP and the decoder
\begin{align}
    \stateApprox=\reconstruction \circ \regression(\params, \timee).
    \label{eq: surrogate}
\end{align}
To adjust the weights~$\Weights=\{\Weights_{\text{enc}}, \Weights_{\text{dec}}, \Weights_{\text{mlp}}\}$ of the networks given some data, we minimize the loss function
\begin{subequations}
    \begin{align}
        \loss
        &= \lossWeights_\text{approx}(\states-\stateApprox)^2 + 
        \lossWeights_\text{rec}(\states-\stateRec)^2 
        \label{eq: loss}
        \\
        &=\underset{\text{Decoder and \MLP}}
            {\underbrace
                {\lossWeights_\text{approx}(\states-\reconstruction(\regression(\params, \timee, \Weights_{\text{mlp}}), \Weights_{\text{dec}}))^2}} 
                \label{eq: loss dec and mlp}
                \\
        &+ \underset{\text{Encoder and Decoder}}
            {\underbrace{
                    \lossWeights_\text{rec}(\states-\reconstruction(\reduction(\states, \Weights_{\text{enc}}), \Weights_{\text{dec}}))^2}} 
                    \label{eq: loss enc and dec}.
    \end{align}
\end{subequations}
The first part of the loss~\cref{eq: loss dec and mlp} ensures that the surrogate captures the system behaviour for given parameters, the second part of the loss~\cref{eq: loss enc and dec} ensures that the autoencoder is able to reconstruct the state from the latent space well.

\subsection{Transfer Learning}\label{subsubsec4}
\begin{figure}
    \centering
    \ifthenelse{\boolean{overleaf}}
    {
        \includegraphics[]{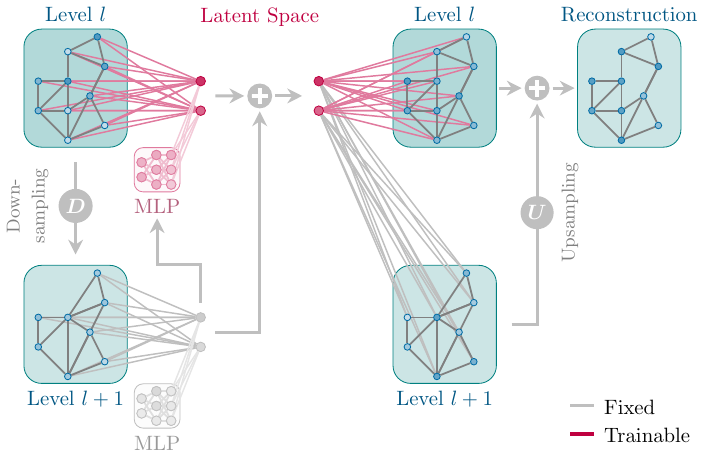}
    }
    {
        \input{fig/methods/Transfer_Learning/transfer_learning.tex}
    }
    \caption{Transfer learning step inside the multi-hierarchical surrogate modeling approach. The weights of the coarse network are fixed, so that the finer surrogate only needs to capture not covered aspects.}
    \label{fig: transfer learning}
\end{figure}

Once a surrogate is found on the coarsest level, the surrogate modeling process can be repeated on the next level of refinement. However, instead of learning everything from scratch, the finer surrogate uses the output of the already trained coarser surrogate.
To transfer the knowledge from one level to another, we connect the finer and the coarse graph representations of the system via down- and upsampling matrices, fix the already trained coarse surrogate, and add its output in the latent space and the reconstruction of the fine surrogate as can be seen in \cref{fig: transfer learning}. 
Consequently, the fine model only needs to capture inaccuracies and non-captured system behavior of the coarse one. The general architecture of the finer decoder, encoder, and \MLP follow the previous definitions. 
The following section explains how to construct the multi-hierarchical model.

\subsubsection{Multi-hierachical Model}\label{subsubsec6}
The starting points for creating the multi-hierarchical model are the differently resolved discretizations of the original system. 
The multi-hierachical modeling approach starts with creating a surrogate~$\surrogate_{\nlevel}$ on the deepest, i.e. coarsest level~$\nlevel$ and just follows the definitions given in~\cref{eq: autoencoder} and~\cref{eq: surrogate} but operates on a downsampled state~$\states_{\nlevel}=\downsampling_{\nlevel} \states$ instead of the original state description. Once the coarse surrogate is created, the surrogate modeling process continues to the next finer level.

The weights~$\Weights_\nlevel$ of the already trained coarse surrogate are fixed to train the finer surrogate~$\surrogate_{\nlevel\text{-}1}$.
The first adjustment compared to the presented standard modeling scheme takes place in the encoding of the system state. Instead of having a single encoder~$\redState_{\nlevel} = \reductionSubscript{\nlevel}(\states_{\nlevel})$ as previously, the latent state is computed as an addition of two encoders  
\begin{align}
    \begin{split}
    \redState_{\nlevel\text{-}1}  
    &= \reductionSubscript{\nlevel\text{-}1}  (\states_{\nlevel\text{-}1}) 
    \\
    &= \reductionSubscript{\nlevel}(\downsampling_{\nlevel\text{-}1}^{\nlevel} \states_{\nlevel\text{-}1} )
        + \reductionSubscript{\nlevel\text{-}1}^{*}(\states_{\nlevel\text{-}1} ) 
        \\
    &= \reductionSubscript{\nlevel}(\states_{\nlevel}) + \reductionSubscript{\nlevel\text{-}1}^{*}(\states_{\nlevel\text{-}1})\\
    &= \redState_{\nlevel} + \reductionSubscript{\nlevel\text{-}1}^{*}(\states_{\nlevel\text{-}1} )
    \end{split}\label{eq: refined encoder}
\end{align}     
with~$\downsampling_{\nlevel\text{-}1}^{\nlevel}$ being the downsampling matrix that maps a state from level~$\nlevel\text{-}1$ to ~$\nlevel$. In this context~$\reductionSubscript{\nlevel}$ is the trained and fixed encoder from the coarse level and~$\reductionSubscript{\nlevel\text{-}1}^*$ is a new trainable encoder. 

A similar approach is chosen to reconstruct the state in the physical space. Therefore, we rely on an addition of the already trained decoder~$\reconstructionSubscript{\nlevel}$ and a trainable new one~$\reconstructionSubscript{\nlevel\text{-}1}^{*}$ resulting in the refined decoder 
\begin{align}
        \stateRec_{\nlevel\text{-}1}
        &= \reconstructionSubscript{\nlevel\text{-}1}(\redState_{\nlevel\text{-}1}) \nonumber\\
        &= \upsampling_{\nlevel}^{\nlevel\text{-}1}\reconstructionSubscript{\nlevel}(\redState_{\nlevel\text{-}1})
            + \reconstructionSubscript{\nlevel\text{-}1}^{*}(\redState_{\nlevel\text{-}1})    
        \label{eq: refined decoder}
        \\
        &= \upsampling_{\nlevel}^{\nlevel\text{-}1}\stateRec_{\nlevel}
            + \reconstructionSubscript{\nlevel\text{-}1}^{*}(\redState_{\nlevel\text{-}1})\nonumber
\end{align}  
where~$\upsampling_{\nlevel}^{\nlevel\text{-}1}$ describes the upsampling matrix from level~$\nlevel$ to~$\nlevel\text{-}1$. 
This static and error-prone upsampling matrix can be replaced with an adaptive learnable upsampling scheme to further minimize the error.
In this work, we decided to use a simple linear fully-connected layer of the form
\begin{align}
    \states_{\nlevel\text{-}1} \approx \samplingNetwork^{\nlevel\text{-}1}_{\nlevel}(\states_{\nlevel}, \Weights_{\text{up}, \nlevel\text{-}1}) \coloneqq \Weights_{\text{up}, \nlevel\text{-}1}^{\setminus\bm{0}}\states_{\nlevel} + \Weights_{\text{up}, \nlevel\text{-}1}^{\bm{0}}
    \label{eq: upsampling layer}
\end{align}
which prooved sufficient in experiments to significantly reduce the upsampling error while still maintaining limited computational effort. The trainable parameters consist of the weights~$\Weights_{\text{up}, \nlevel\text{-}1}^{\setminus\bm{0}}$ and the bias~$\Weights_{\text{up}, \nlevel\text{-}1}^{\bm{0}}$.
Replacing the former upsampling matrix with~\cref{eq: upsampling layer} in Eq.~\cref{eq: refined decoder} leads to the decoder formulation we are using in this work
\begin{align}
        \reconstructionSubscript{\nlevel\text{-}1}(\redState_{\nlevel\text{-}1}) 
        = \samplingNetwork^{\nlevel\text{-}1}_{\nlevel}(\reconstructionSubscript{\nlevel}(\redState_{\nlevel\text{-}1}))
            + \reconstructionSubscript{\nlevel\text{-}1}^{*}(\redState_{\nlevel\text{-}1}).  
        \label{eq: refined decoder with adaptive upsampling}
\end{align}   

The refined multilayer perceptron on the contrary uses the previous one's output as additional input 
\begin{align}
    \redStateApprox_{\nlevel\text{-}1} 
    &= \regression_{\nlevel\text{-}1}(\params, \timee)
    = \regression^{*}_{\nlevel\text{-}1}(\params, \timee, \regression_{\nlevel}(\params, \timee))
    = \regression^{*}_{\nlevel\text{-}1}(\params, \timee, \redStateApprox_{\nlevel}(\params, \timee)).
    \label{eq: refined mlp}
\end{align} 
To create the next finer surrogate model~$\surrogate_{\nlevel\text{-}2}$, the same procedure is repeated but this time 
with~$\surrogate_{\nlevel\text{-}1}$ serving as coarse model. 

To enable a comparison among all levels, it is of interest to transform the approximations of the surrogate models back into the original finely discretized state space. 
For the upsampling of the coarse approximation to the original discretization, the static upsampling matrices~$\upsampling_{\layerIterator}^{0},\ 1\leq\layerIterator\leq\nlevel$ are used 
\begin{align*}
    \stateApprox_{\layerIterator}^{0} = \upsampling_{\layerIterator}^{0}(\stateApprox_{\layerIterator}).
\end{align*}
The presented framework offers many adjustments to adapt it to one's own needs, and the presented decision choices represent only one suitable configuration. Some specific variations are mentioned in the following.

\subsection{Alternative Architectures}
\label{sec:alternative architectures}
One point at which adjustments can be made to the presented architecture are the refined versions of the encoder \cref{eq: refined encoder}, decoder \cref{eq: refined decoder with adaptive upsampling}, and \MLP~\cref{eq: refined mlp}. 
Instead of having additive transfer learning for the encoder, the results from several layers could be concatenated, which lead to a higher latent dimension. This drawback, along with the absence of a performance boost in numerical experiments, has led to the abandonment of this idea.
Furthermore, adding output of the coarse decoder to the input for the fine one, as we did for the \MLP, would significantly increase in the number of input dimensions and is generally difficult for graph convolutions. 

Another design modification can be made in the adaptive upsampling \cref{eq: upsampling layer}. Either by replacing the proposed upsampling mapping, e.g. with another type of layer, by optimizing the sparse upsampling matrix~$\upsampling_{\layerIterator}^{m}$, or by carefully selecting the nodes that require a refinement.

\subsubsection{Adaptive Refinement}\label{subsubsec5}
The idea of adaptive refinement is that only areas of certain interest or high error are refined in the surrogate modeling process. That means that only the coarsest surrogate is trained on the precomputed discretizations. All subsequent finer levels will only use the refined version for areas where it is desired. 

\begin{figure}
    \centering
    \ifthenelse{\boolean{overleaf}}
    {
        \includegraphics[]{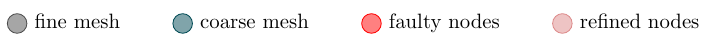}
    }
    {  
        \begin{tikzpicture}
            \matrix [below right] at (current bounding box.north east) {
            \node [circle, fill=fineColor!50!white, draw=fineColor] (l1)  {}; 
            \node [right=0mm of l1, anchor=west] (l1) {fine mesh\phantom{y}}; 
            \node [circle, anchor=center, fill=coarseColor!50!white, draw=coarseColor] (l2) at ($(l1.east) + (0.75, 0)$) {}; 
            \node [right=0mm of l2.east, anchor=west] (l2) {coarse mesh\phantom{y}}; 
            \node [circle, anchor=center, fill=red!50!white, draw=red] (l3) at ($(l2.east) + (.75, 0)$) {}; 
            \node [right=0mm of l3, anchor=west] (l3) {faulty nodes\phantom{y}}; 
            \node [circle, anchor=center, fill=refinedColor!50!white, draw=refinedColor] (l4) at ($(l3.east) + (.75, 0)$) {}; 
            \node [right=0mm of l4, anchor=west] (l4) {refined nodes\phantom{y}}; 
            \\
            };
        \end{tikzpicture}
    }
	\begin{subfigure}[t]{0.3\textwidth}
        \includegraphics[page=1, width=\textwidth]{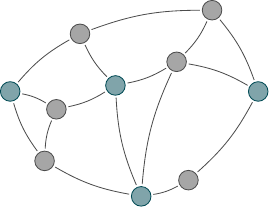}
        \caption{Fine mesh}
    \end{subfigure}
    \hfill
	\begin{subfigure}[t]{0.3\textwidth}
        \includegraphics[page=2, width=\textwidth]{fig/faulty_node.pdf}
        \caption{Coarse mesh}
    \end{subfigure}     
    \hfill
    \begin{subfigure}[t]{0.3\textwidth}
        \includegraphics[page=3, width=\textwidth]{fig/faulty_node.pdf}
        \caption{Refined mesh}
    \end{subfigure}
    \caption{Adaptive selection of faulty nodes, i.e. of nodes with errors above a defined tolerance measured on validation data, to refine meshes in areas of high error.}
    \label{fig: faulty nodes}
\end{figure}

A possible data-based approach to select those areas is to calculate a suitable error of the coarse surrogate on a validation dataset and then chose those nodes which have the highest error or penalize an error threshold. We refer to them as faulty nodes. For the selected nodes,  all neighboring nodes in the next finer graph are added, see \cref{fig: faulty nodes}. In the next step, the adjacency matrix defining the resulting graph and suitable up- and downsampling matrices need to be computed. 
While this approach gives special consideration to areas of interest, e.g., areas with high variability, and is appealing due to the smaller resulting models and the possibility to include error tolerances (for validation data), it also leads to a complicated framework. In addition, it introduces several design decisions, such as how to define the neighborhood of a refined node. Furthermore, no notable performance boost could be observed in our numerical experiments compared to our vanilla version, and the reduction in computational costs is minimal.
Consequently, we only present the results of the more comprehensive vanilla approach in the following.

\subsubsection{Unified Latent Representation}
The multi-hierarchical representations of the original system not only enable multiple surrogate models to be trained one after another but can also be used for simultaneous training. In such an approach, the different latent representations for every level~$\redState_\layerIterator, \ 1\leq\layerIterator\leq\nlevel$ could be exchanged for one unified description so that the model distinction only takes place in the decoders. This advantage is bought by the disadvantage of not being able to stop the refinement at any point and to learn global behavior very easily and fast in a simple representation.

\section{Results \& Discussion}\label{sec: results}
To highlight the performance of the presented approach, we present numerical results for the aforementioned simplified crash simulation. The created surrogates are rated regarding their training phase, approximation quality in the coarse and the original representation, and their computing time. All following results, with exception of the finite element simulations were produced on an Apple M1 Max with a 10-Core CPU, 24-Core GPU, and 64 GB of RAM.
To compare the proposed framework with more classic approaches, we generate surrogate models that follow the description of \cref{sec:surrogate} but operate directly on the original (not downsampled) data and use either proper orthogonal decomposition, fully connected autoencoder or a graph convolutional autoencoder for the reduction step. We refer to them as PODNN, AENN, and GAENN, whereas the surrogates using the multi-hierarchical approach with graph convolutional autoencoders on the different levels are referred to as MH1, MH2, and MH3 (from finest to coarsest surrogate). The chosen architectures are listed in \cref{tab:architecture}.

The \MH encoders consist of several graph convolutional layers with ELU activation functions. Each graph convolution maintains the signal dimension~$\nNodes$ but changes the number of channels~$\nChannels$. The graph convolutions are followed by a dense layer with linear activation function to map the input to the latent dimension. The decoder follows the same architecture in reverse order.
All dimensionality reduction networks~(POD, AE, GAE, MH1, MH2, MH3) are combined with similar MLPs to predict the latent state based on the simulation parameters. Each one consists of several fully-connected layers with ELU activation function and a final dense layer with linear activation function. The surrogates are trained for 1500 epochs, and the weights with the lowest total loss~\cref{eq: loss} 
are then used for subsequent predictions.
Another noteworthy aspect showcased in \cref{tab:architecture} is that the graph convolutional networks possess much fewer parameters than a comparable multi-layer fully-connected networked architecture.
\begin{table}
	\setlength{\tabcolsep}{3pt}
    \small
	\centering
	\caption{Model Architectures. The transitions between the network states are described by the used activation function (elu, linear) and layer type (fully-connected (fc) and Chebyshev graph convolutional layer (gcn) of order~$\chebOrder=3$). For the autoencoder models, only the structure of the encoder is given, whereas the decoder follows the same architecture but in reverse order.}
	\label{tab:architecture}
	\npdecimalsign{.}
	\nprounddigits{3}
	\begin{tabularx}{\textwidth}{@{} llX @{}}
		\toprule
		\textbf{Model} & \textbf{Network State Sizes} & \textbf{Number of Parameters} \\
          \midrule
		POD &  \footnotesize{$27942 \overset{\text{linear}}{\to} 4$}& 111\,768 \\
        AE & \footnotesize{$27942 \overset{\text{fc,elu}}{\to} 200 \overset{\text{fc,elu}}{\to} 80 \overset{\text{fc,linear}}{\to} 4$} & 11\,246\,854\\
        GAE 
            &  \footnotesize{$9314\times3 \overset{\text{gcn,elu}}{\to} 9314\times6 \overset{\text{gcn,elu}}{\to} 9314\times12$  $\overset{\text{gcn,elu}}{\to}  9314\times24  \overset{\text{fc,linear}}{\to}  4$} 
            & 2\,023\,067\\
        MH1 
            &   \footnotesize{$1165\times3 \overset{\text{gcn,elu}}{\to} 1165\times6 \overset{\text{gcn,elu}}{\to} 1165\times12$ $\overset{\text{gcn,elu}}{\to} 1165\times24 \overset{\text{fc,linear}}{\to} 4$ }
            & 253\,987\\
        MH2 
            &   \footnotesize{$292\times3 \overset{\text{gcn,elu}}{\to} 292\times6 \overset{\text{gcn,elu}}{\to} 292\times12$ $\overset{\text{gcn,elu}}{\to} 292\times24 \overset{\text{fc,linear}}{\to} 4$} 
            & 65\,419\\
        MH3 
            &  \footnotesize{$73\times3 \overset{\text{gcn,elu}}{\to} 73\times6 \overset{\text{gcn,elu}}{\to} 73\times12$ $\overset{\text{gcn,elu}}{\to} 73\times24 \overset{\text{fc,linear}}{\to} 4$} 
            & 18\,115  \\
        MLP  &  \footnotesize{$4 \overset{\text{fc,elu}}{\to} 64 \overset{\text{fc,elu}}{\to} 64 \overset{\text{fc,elu}}{\to} 64 \overset{\text{fc,linear}}{\to} 4 $} & 8\,900 \\
		\bottomrule
	\end{tabularx}
	\npnoround
\end{table}

\subsection{Numerical Example of a Racing Kart}

For the task of creating a surrogate for the crash model, we are interested in approximating the kart's displacements~$\disps(\timee, \params)\in\Rdim^{\nNodes \times 3}$ based on the simulation parameter~$\params$ and the time~$\timee$. As all simulations start from the same initial condition, it is neglected in the modeling of the surrogate. The displacement of the~$\nodeIteratorA$-th node of the~$\simIterator$-th simulation at time~$\timee$ is denoted by~$\disps_{\nodeIteratorA}^{\simIterator,}(\timee)=\left[\disp_{\nodeIteratorA}^{\simIterator,x}(\timee), \disp_{\nodeIteratorA}^{\simIterator,y}(\timee), \disp_{\nodeIteratorA}^{\simIterator,z}(\timee)\right]\in\Rdim^{3}$, where the superscripts~$x,\ y,\ z$ represent the corresponding coordinate direction.   
In total $\nSims=128$ parameter combinations were sampled using Halton sequences. From the resulting high-fidelity simulation results, $\nSims^{\text{train}}=96$ are used for training and~$\nSims^{\text{test}}=32$  to test the surrogates.

For the graph convolutional based surrogates, the displacements directly serve as state, i.e.~$\states\coloneqq\disps$, whereas they need to be vectorized for the other ones, i.e.~$\states\coloneqq \left[\disps_{1}^{\simIterator,}, \dots, \disps_{\nNodes}^{\simIterator}\right]^\transpose \in\Rdim^{3\nNodes}$.
Consequently, the dataset
\begin{align}
    \dataSet \coloneqq
    \begin{Bmatrix}
        \begin{bmatrix}
            \timee_0 &\dots& \timee_{\nTime-1}\\
            \params_\simIterator &\dots& \params_\simIterator\\
        \end{bmatrix},
        \begin{bmatrix}
            \states^{\simIterator}(\timee_0) &\dots& \states^{\simIterator}(\timee_{\nTime-1}) 
        \end{bmatrix}
    \end{Bmatrix}_{\simIterator=1}^{\nSims^{\text{train}}},
\end{align} 
contains the time~$\timee$ and the parameters~$\params$ as input for the \MLP and the corresponding displacements as target values.
All simulation results as well as the kart's source files are published and freely available under~\cite{darus-3789_2023}.
With the data created, the surrogate modeling process can be started.

\subsection{Training Comparison between Fine and Coarse Models}
\begin{figure}[t!]
    \centering
    \begin{subfigure}[t]{0.32\textwidth}    
        \ifthenelse{\boolean{overleaf}}
        {
            \includegraphics[]{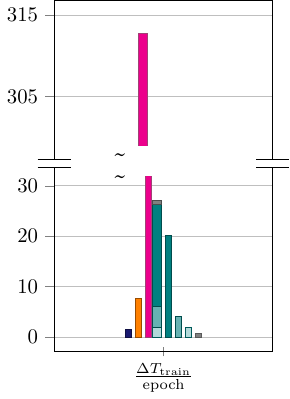}
        }
        {
            \input{fig/results/kart/training_time/kart_training_time_with_broken_y_axis.tex}
        }   
        \caption{Training time per epoch using 9696 samples and a batch size of~32.}
        \label{fig:training_time_kart}
    \end{subfigure}
    \hfill
    \begin{subfigure}[t]{0.64\textwidth}   
        \ifthenelse{\boolean{overleaf}}
        {
            \includegraphics[]{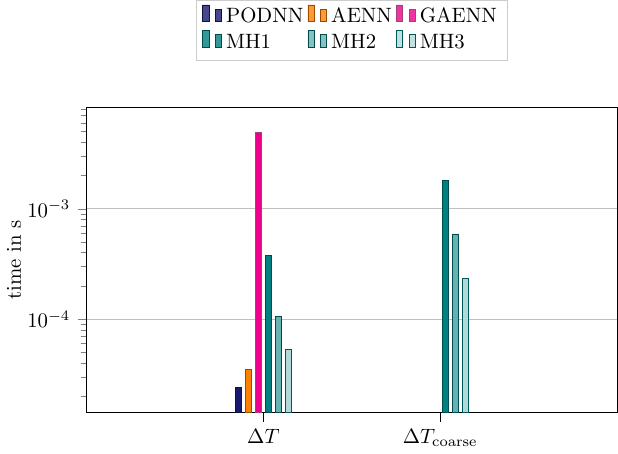}
        }
        {
            \input{fig/results/kart/training_time/kart_computing_time.tex}
        }    
        \vspace{-\baselineskip}
        \caption{Averaged computing time per prediction on full model (with upsampling) and on the coarse levels (without upsampling)}
        \label{fig:computing_time_kart}
    \end{subfigure}
    \caption{Computational time required to train the models~(a) and to compute a prediction~(b) for the kart model on the fine and the coarse mesh using the proposed multi-hierarchical approach and standard approaches.}
    \label{fig:times_kart}
\end{figure}

Before we evaluate the actual performance of the surrogate models, let's first take a glance into the training phase.
For an overview of the data on which the different surrogate models are trained and which variables are optimized in this process, please refer to \cref{tab:parameters}.
A drawback of the graph convolution is the associated computational cost that among others arises from the recursive computation of the Chebyshev polynomials. Consequently, the time to train a graph convolutional surrogate model on the full model significantly exceeds the training time ofa classic fully-connected autoencoder as shown in \cref{fig:training_time_kart}. If the surrogate is created using the multi-hierarchical approach on the contrary the tide turns. The training time is reduced to such an extent that the model operating on the coarsest representations trains even faster than the classical autoencoder on the full model. Even when adding up the training time of all three levels used for the kart example, the time is still in a comparable order of magnitude and is more than ten times faster than the GAENN on the full model. 

\begin{table}
	\setlength{\tabcolsep}{3pt}
	\centering
	\caption{Overview of surrogate models showing which data is used and which parameters are optimized.}
	\label{tab:parameters}
	\npdecimalsign{.}
	\nprounddigits{3}
	\begin{tabularx}{\linewidth}{@{}>{\bfseries}l@{\hspace{1.em}}X@{\hspace{1em}}l@{\hspace{1.em}}l@{\hspace{1.em}}@{}}
		\toprule
            \textbf{Model} 
            & \textbf{Reduction Algorithm}
            & \textbf{Mesh data}
            & \textbf{Trainable Parameters}
            \\
          \midrule
		PODNN 
            & {Proper orthogonal decomposition} 
            & $\states$ 
            & $\Weights_{\text{mlp}}^{\text{pod}}$ 
            \\
        AENN 
            & {Fully-connected autoencoder} 
            & $\states$ 
            & $\Weights_{\text{mlp}}^{\text{ae}}, \Weights_{\text{enc}}^{\text{ae}}, \Weights_{\text{dec}}^{\text{ae}}$ 
            \\
        GAENN
            & {Graph convolutional autoencoder} 
            & $\states$ 
            & $\Weights_{\text{mlp}}^{\text{gae}}, \Weights_{\text{enc}}^{\text{gae}}, \Weights_{\text{dec}}^{\text{gae}}$ 
            \\
        MH1 
            & {Multi-hierarchical graph conv. autoencoder} 
            & $\states_{1}=\downsampling_1\states$ 
            & \makecell[l]{$\Weights_{\text{mlp}}^{\text{mh1}}, \Weights_{\text{enc}}^{\text{mh1}},$ $\Weights_{\text{dec}}^{\text{mh1}}, \Weights_{\text{up},1}^{\text{mh1}}$} 
            \\
        MH2 
            & {Multi-hierarchical graph conv. autoencoder} 
            & $\states_{2}=\downsampling_2\states$ 
            & \makecell[l]{$\Weights_{\text{mlp}}^{\text{mh2}}, \Weights_{\text{enc}}^{\text{mh2}},$ $\Weights_{\text{dec}}^{\text{mh2}}, \Weights_{\text{up},2}^{\text{mh2}}$}
            \\
        MH3 
            & {Graph convolutional autoencoder} 
            & $\states_{3}=\downsampling_3\states$ 
            & $\Weights_{\text{mlp}}^{\text{mh3}}, \Weights_{\text{enc}}^{\text{mh3}}, \Weights_{\text{dec}}^{\text{mh3}}$ 
            \\
		\bottomrule
	\end{tabularx}
	\npnoround
\end{table}

Considering the computing time required~$\compTime$ for one prediction of the surrogates, a similar picture emerges as depicted in \cref{fig:computing_time_kart}. The GAENN requires by far the most time but our approach can substantially mitigate this effect.  A surrogate that just uses \POD for the dimensionality reduction outperforms the other models as the computation of the reconstruction to the fine physical space only requires one matrix multiplication. 
Regarding the \MH approach, the training and computational time logically increases with every level as the degree of resolution rises. Noteworthy in this context is that the difference in time to receive a prediction for the fine original representation is not much higher than that one of the coarse representations. This is owed to the fact that the upsampling follows \cref{eq: upsampling} and consequently only requires a sparse matrix multiplication. 

\begin{figure}[t!]
    \centering
    \begin{subfigure}[t]{0.32\textwidth}  
        \ifthenelse{\boolean{overleaf}}
        {
            \includegraphics[]{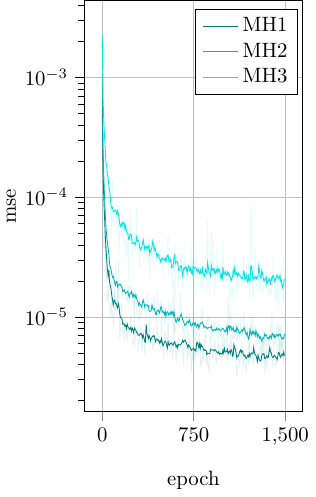}
        }
        {
            \input{fig/results/kart/training_history/kart_loss_training_history.tex}
        }
        \caption{Loss~\cref{eq: loss}
        }
        \label{fig: global loss}
    \end{subfigure}
    \hfill
    \begin{subfigure}[t]{0.32\textwidth}  
        \ifthenelse{\boolean{overleaf}}
        {
            \includegraphics[]{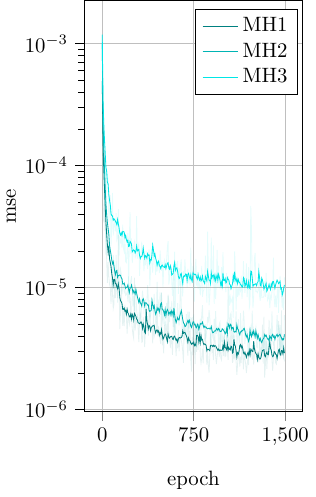}
        }
        {
            \input{fig/results/kart/training_history/kart_approx_training_history.tex}
        }
        \caption{Approx. loss~\cref{eq: loss dec and mlp}
        }
        \label{fig: approx loss}
    \end{subfigure}
    \hfill
    \begin{subfigure}[t]{0.32\textwidth}  
        \ifthenelse{\boolean{overleaf}}
        {
            \includegraphics[]{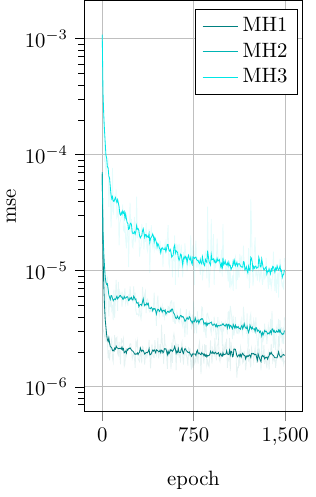}
        }
        {
            \input{fig/results/kart/training_history/kart_rec_training_history.tex}
        }
        \caption{Reconstr. loss~\cref{eq: loss enc and dec}
        }
        \label{fig: rec loss}
    \end{subfigure}
    \caption{Training history for the overall loss~(\subref{fig: global loss}), the approximation loss~(\subref{fig: approx loss}), and the reconstruction loss~(\subref{fig: rec loss}) of the \MH models for the racing kart example. The loss functions are smoothed for a clearer representation, whereas the true values are drawn with transparency in the background. It can be seen that the loss decreases significantly with each refinement compared to the previous coarser level.}
    \label{fig: loss}
\end{figure}

In addition to computation times, training the \MH models provides additional insights. 
For an evaluation of the transfer learning, we consider the progression of the loss during training in \cref{fig: loss}. It becomes apparent that all losses drop significantly lower with each refinement of the models. Consequently, the saved information of the coarser models helps the finer ones to improve their performance and avoids that already known structures have to be learned twice. Especially, the reconstruction benefits greatly from the transfer learning, see \cref{fig: rec loss}, but the overall approximation gets better with each level as well, see \cref{fig: approx loss}. 

\subsection{Evaluation on Coarse Levels}
In the final comparison of the surrogate models, the performance metrics are always measured in the original discretization of the model so that a comparison between the \MH models and models that are trained on the original data directly is possible. 
However, there are two reasons why it is worth considering the performance in the coarse discretizations prior to this final consideration: On the one hand, because the \MH models are intended to be evaluated in the coarse representation and on the other hand because in this way the error induced by the upsampling into the original space does not appear in the evaluation. Furthermore, we can compare how the graph convolutional \MH models perform in comparison to standard surrogate models on each level and thus justify their use.

For a comparison of the performance we utilize the averaged Euclidean distance between the nodes of the reference \FE simulation and their approximation
\begin{align*}
    \error_{2,\simIterator}(t) = \frac{1}{\nNodes}\sum_{\nodeIteratorA=1}^{\nNodes} \lvert \disp_{\nodeIteratorA, \simIterator}(\timee) - \approximate{\disp}_{\nodeIteratorA, \simIterator}(\timee) \rvert_2 
\end{align*} 
at time~$\timee$ of the~$\simIterator$-th simulation.
Moreover, its mean value over the time
\begin{align*}
    \hat{\error}_{2,\simIterator} = \mean{\timee\in\timeSpace} \error_{2,\simIterator}(t)
\end{align*} 
and the mean value over time and all test simulations
\begin{align*}
    \eTwoMean = \mean{\simIterator\in [1, \dots, \nSims^{\text{test}} ]} \hat{\error}_{2,\simIterator}
\end{align*} 
are used to represent the approximation quality for the complete test data.

\begin{figure}[t!]
    \centering
    \ifthenelse{\boolean{overleaf}}
    {
        \includegraphics[]{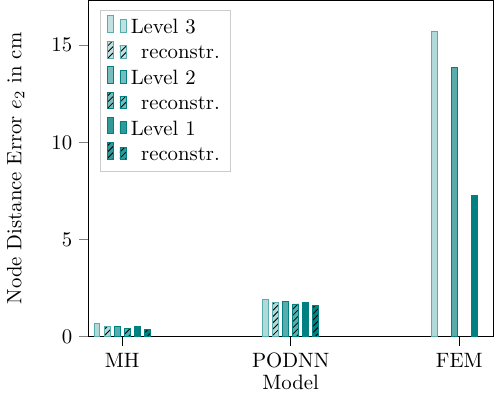}
    }
    {
        \input{fig/results/kart/kart_error_level/kart_error_level_save.tex}
    }
    \caption{Mean node error of the kart model on different coarse meshes using the proposed multi-hierarchical approach and a standard PODNN approach. The errors is once measured for plain reconstruction of the state and once for the reconstruction of the approximated state. Note that the FEM errors are calculated for a slightly different reference model for technical reasons (connection of point masses to model).}
    \label{fig: error on coarse levels}
\end{figure}

The first investigation on the coarse meshes is conducted to emphasize the hypothesis stated at the beginning; that the degrees of freedom result from the necessity of the modeling method. Therefore, finite element models that only differ in their discretization are generated, and the same scenario is simulated for all of them. As shown in \cref{fig: error on coarse levels}, the node distance between the coarse \FE models and the reference position of the corresponding selected nodes in the original mesh is far apart. The results reveal qualitatively different dynamic behavior and confirm the need for a fine resolution using the finite element method. Note that the coarse \FE models are only produced with the presented downsampling approach and not with a proper mesh simplification method for finite element models. Nevertheless, the results show of how much conventional methods rely on a fine resolution. 

Considering the node distance error of the \MH models and PODNN surrogates on the different discritizations, two essential points are noteworthy. On the one hand, the all \MH models outperform the POD-based surrogate. Accordingly, the graph convolutional architecture works well on the coarsest representation and justifies its use compared to other similarly applicable architectures. On the other hand, the error decreases with every additional level for the \MH models. In an error view of the original fine resolution of the kart, this is not surprising, since the upsamling error decreases with each level. On the coarse discretizations, however, this clearly indicates that transfer learning helps to lower the error at each level.

\begin{figure}[t!]
    \centering
	\begin{subfigure}[t]{\textwidth}
        \ifthenelse{\boolean{overleaf}}
        {
            \includegraphics[]{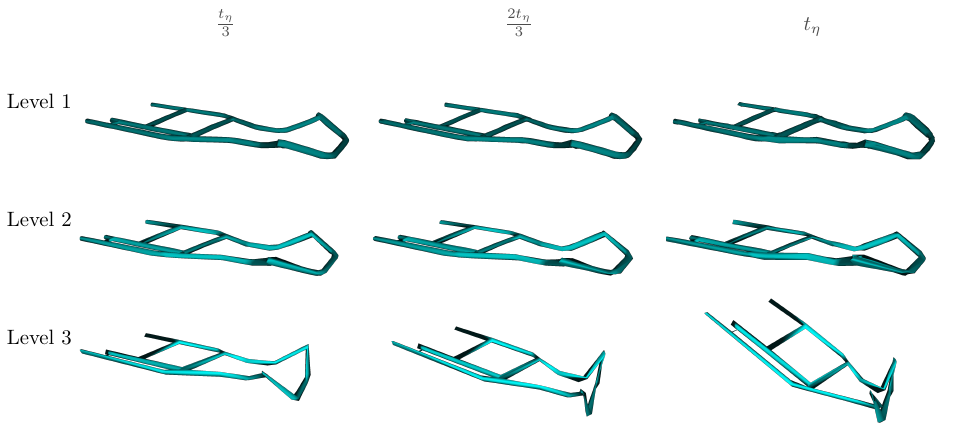}
        }
        {
            \begin{tikzpicture}
            \def\z{0}
                \node [] at (0, 0) (level10) {Level 1};
                \node [] at (0, -2) (level20) {Level 2};
                \node [] at (0, -4) (level30) {Level 3};
                \foreach \x/\y in {1/$\frac{\timee_\nTime}{3}$, 2/$\frac{2\timee_\nTime}{3}$, 3/$\timee_\nTime$}{
                    \node [inner sep=0pt, anchor=west, 
                        label={[align=center,yshift=-.75cm, text=black!70!white,label distance=0cm]\textbf{\y}}
                        ] 
                        (level1\x) at ($(level1\z.east)+(-0.0cm,00cm)$)
                        {\includegraphics[trim=14cm 12cm 12cm 0cm, clip, width=.3\textwidth]{kart_contribution_level0_compl_\x.png}};
                    \node [inner sep=0pt, anchor=west
                        ] 
                        (level2\x) at ($(level2\z.east)+(-0.0cm,00cm)$)
                        {\includegraphics[trim=14cm 12cm 12cm 0cm, clip, width=.3\textwidth]{kart_contribution_level1_compl_\x.png}};
                    \node [inner sep=0pt, anchor=west
                        ] 
                        (level3\x) at ($(level3\z.east)+(-0.0cm,00cm)$)
                        {\includegraphics[trim=14cm 12cm 12cm 0cm, clip, width=.3\textwidth]{kart_contribution_level2_compl_\x.png}};
                    \global\let\z=\x
                }
            \end{tikzpicture}%
        }
    \end{subfigure}
    \caption{Learned behavior at the different levels. Global behavior is learned at the coarsest, i.e., deepest level. The learned behavior is then transferred to the finer levels, where only minor adjustments are made.
    }
    \label{fig: level contributions}
\end{figure}

It is important to emphasize which dynamic effects are learned at which level. 
For a visual illustration, the learned behavior at each level for an example simulation is given in \cref{fig: level contributions}.
It showcases the models approximation subtracting the already existing prediction of the coarse levels, i.e.
\begin{align*}
    \stateApprox_{\nlevel\text{-}1}^{0} - \stateApprox_{\nlevel}^{0}
        &= \upsampling_{\layerIterator\text{-}1}^{0}
            \reconstructionSubscript{\nlevel\text{-}1}(\regression_{\nlevel\text{-}1}(\params, \timee)) 
        - \upsampling_{\layerIterator}^{0}\upsampling_{\nlevel}^{\nlevel\text{-}1}
            \reconstructionSubscript{\nlevel}(\regression_{\nlevel}(\params, \timee))
    .
\end{align*}
Clearly, the global dynamic behavior is already captured in the coarsest surrogate and the finer ones only learn minor adjustments to compensate for local errors.

\subsection{Approximation Quality}
\begin{figure}[t!]
    \centering
    \ifthenelse{\boolean{overleaf}}
    {
        \includegraphics[]{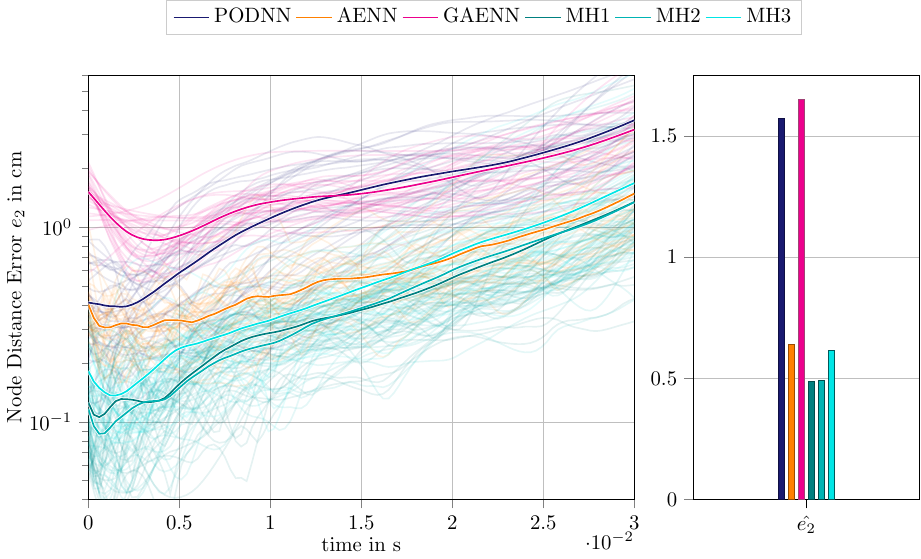}
    }
    {
        \input{fig/results/kart/kart_error_over_time/kart_error_over_time_nice.tex}
    }
    \caption{Mean node error over time for all test simulations achieved by different surrogate models. The individual test simulations are drawn transparently, while the mean value of all test simulations is shown opaquely.}
    \label{fig: node error over time}
\end{figure}

In a final comparison, we validate the surrogates' performance in the original model discretization. We refer to \cref{fig: node error over time}, where the averaged node distance error over time is shown for the different models. Interestingly, the graph convolutional autoencoder-based surrogate without multi-hierarchical structure fails to capture the dynamics and consequently has the largest error. The surrogate using linear reduction in form of the \POD struggles to approximate the intervals of high dynamics as only~$\redStateDim=4$ reduced basis vectors are not expressive enough to describe all complex deformations occurring in the simulations.

The AENN surrogate model relying on a classic autoencoder already shows promising results indicating the benefits of a nonlinear dimensionality reduction. Nevertheless, even the coarsest \MH model beats it in average, and the error decreases with each subsequent finer level. 
The most important performance indicators are summarized in \cref{tab:perf} to provide the main results at a glance.

\begin{table}
	\setlength{\tabcolsep}{3pt}
	\centering
	\caption{Performance measurements. The time specification in brackets indicate the summed up training time over all levels in case of the \MH models and the training time for the \MLP in case of PODNN.}
	\label{tab:perf}
	\npdecimalsign{.}
	\nprounddigits{2}
	\begin{tabular}[c]{l l l l l l l}
		\toprule
		&\textbf{PODNN}& \textbf{AENN} & \textbf{GAENN} & \textbf{MH1} & \textbf{MH2} & \textbf{MH3}\\
		\midrule
		$\frac{\compTime_{\text{train}}}{epoch}$ in s
			&-~($\numprint{1.539057379}$)
			&$\numprint{7.749005098}$ 
			&$\numprint{312.7500264}$ 
			&$\numprint{20.21926474}$~($\numprint{27.0783241}$)
			&$\numprint{4.095964556}$~($\numprint{6.27019866}$)
			&$\textbf{\numprint{1.956770506}}$ \\
        $\compTime$ in ms
			&$\textbf{\numprint{0.0241689457751736}}$ 
			&$\numprint{0.0351850349124115}$ 
			&$\numprint{0.90708036883042}$ 
			&$\numprint{0.380754618361444}$    
			&$\numprint{0.105726468090963}$ 
			&$\numprint{0.0530828786368417}$\\
        $\eTwo$ in cm 
			&$\numprint{1.573408768}$ 
			&${\numprint{0.642048288}}$ 
			&${\numprint{1.653678343}}$ 
			&$\textbf{\numprint{0.486192228}}$  
			&${\numprint{0.490680322}}$ 
			&${\numprint{0.616193549}}$ \\
		\midrule
	\end{tabular} 
	\npnoround
\end{table}

\subsection{Discussion}
Our results show that the proposed multi-hierarchical surrogate modeling scheme is suitable for creating various reduced order models for crashworthiness applications. We captured the transient dynamics, including massive plastic deformations of the considered kart's frame resulting from nonlinear contact under multiple parameter dependencies. 
In particular, our method outperforms standard approaches regarding accuracy while still maintaining competitive computational costs and possessing less parameters. Moreover, the \MH surrogates can directly operate on the coarse (less memory-demanding) representations of the system that are still visually interpretable making them suited for graphical application in hardware-restricted use cases. Nevertheless, their predictions can be lifted into the original system description without adding much computational effort by simple sparse matrix multiplication. We could show that the global dynamic behavior occurring in the investigated crash scenario is already captured in the coarsest surrogate and that the finer ones only need to learn microscale effects. Along with this, we were able to determine that the surrogates accuracy increases with each refinement even in the coarse domains. This effect is also reflected in the course of the loss during the networks' training phase where the loss dropped significantly faster for the finer models. Those observations lead to the conclusion that the transfer learning helps the models to converge closer to the reference solution. 
Furthermore, the proposed architecture offers multiple points for adjustments and extensions as stated in \cref{{sec:alternative architectures}}. 
However, those benefits are gained at the expense of a few disadvantages and limitations.

Our approach requires knowledge about the internal (geometrical) structure of a given system. Consequently, data alone is not enough. Furthermore, the model simplification is performed based on the spatial properties of a given system so that other quantities of interest might be lost or must be acknowledged in the mesh simplification process. This process itself adds computational effort to the offline phase, which is negligible compared to the training effort for the networks. 
Additionally, many new design choices and hyperparameters are added by the multi-hierarchical architecture and the use of graph convolutions. This complicates the surrogate modeling process compared to more straightforward approaches like \POD in combination with neural networks. However, even without extensive fine tuning the \MH models are able to beat the conventional methods. An interesting future research direction is the consideration of low-fidelity data to improve the surrogate models. In our current approach, all hierarchical models are derived from the high-fidelity data only. However, it is considerable to use low-fidelity \FE models based on the coarse meshes and incorporate their results into the surrogate models, similar to multi-fidelity approaches that use cheap low-fidelity models to improve high-fidelity predictions~\cite{Kast2020, Conti2023} and learn the resulting residual~\cite{Demo2023}, for example. 
\enlargethispage{\baselineskip}

Another decision worth discussing is the choice of graph convolutions for dimensionality reduction. The multi-hierarchical framework itself works with arbitrary data-driven reduction methods and consequently the \GCNNs can be replaced with other methods as well. Nevertheless, as the mesh simplification already operates on graphs, it is an obvious choice to use this structure in the data to gain benefits. As shown in the results, the graph convolutional based surrogate on the coarse mesh beats a linear reduction technique by far, even when, in this coarse representation, no transfer learning takes place. Furthermore, the graph convolutions are an architecture that can benefit a lot from the multi-hierarchical approach. The computational time savings have a much greater impact as the used convolutions are computationally expensive per se. As the convolutions use parameter sharing along the filters, the networks using them require less trainable parameters but still reveal a better expressiveness of the data within our framework. Interestingly, a graph convolutional-based surrogate operating on the original fine mesh failed to capture the system adequately which may be caused by oversmoothing issues~\cite{Chen2020, Rusch2023}, the difficulty to transport information over distant nodes in such a fine mesh~\cite{Alon2020}, and the spectral bias~\cite{RahamanEtAl2019}. Consequently, the \MH approach not only facilitates a successful learning process but makes it possible in the first place.

\section{Conclusion}\label{sec: conclusion}
In this paper, we derived a structured surrogate modeling scheme producing efficient yet accurate models for a crash simulation system despite its complexity and inaccessible source code. The surrogates require only as many parameters as other linear state-of-the-art counterparts while even outperforming conventional nonlinear data-driven competitors.

To achieve this, our scheme operates on various representations of the crash simulation model with different resolutions instead of relying on a single high-resolution discretization.
This naturally facilitates the approximation of multiscale effects as global dynamics can be learned on coarse resolutions, while microscale dynamics are captured on finer versions. Moreover, we use low-resolution approximations to ease the learning process and improve accuracy of mid- and fine-resolution approximators by transferring knowledge across levels so that finer models only need to capture residuals. Sparse matrix multiplications or adaptive upsampling networks are used to switch between resolutions. 

The surrogates on a single level are built of graph convolutional autoencoders for discovery of suitable low-dimensional representations of the data and fully connected neural networks that cover the parameter-dependent latent dynamics. In doing so, the resulting surrogate models have a satisfying accuracy despite the comparably low number of parameters. 
The hierarchical approach also speeds up the learning process for the graph convolutional surrogates as it eliminates the need to work with the original fine resolution data and creates multiple models with varying memory and computational demands, all operating in a visually and physically interpretable domain. 

However, the involved mesh simplification process is based on spatial criteria, and thus, other information may be lost in the process. Moreover, similar to other nonlinear reduction techniques, it shows its advantages, especially when the system is reduced to its intrinsic size. For large latent spaces, conventional linear methods can still achieve competitive results. Another limitation it shares with data-driven reduced order models is the lack of extrapolation quality. 

In order to continue this promising path, more recent graph convolutional architectures can be used and all hierarchical models can be covered with a single latent variable. Furthermore, the current framework is only based on expensive high-fidelity data. Accordingly, low-fidelity data, which may come directly from coarser discretizations, can also be embedded in the future.

\subsubsection*{Acknowledgments}
Funded by Deutsche Forschungsgemeinschaft (DFG, German Research Foundation) under Germany's Excellence Strategy - EXC 2075 - 390740016. We acknowledge the support by the Stuttgart Center for Simulation Science (SimTech). 

Furthermore, the authors would like to thank the Ministry of Science, Research and Arts of the Federal State of Baden-W\"urttemberg for the financial support within the InnovationsCampus Future Mobility. 

We also wish to acknowledge the support of the National Science Foundation AI Institute in Dynamic Systems grant 2112085. 

\subsubsection*{Authors' contributions}
J.K.: conceptualization, data curation, investigation, methodology, software, validation, visualization, writing—original draft, writing-review and editing. 

J.F.: conceptualization, funding acquisition, methodology, project administration, supervision, writing-review and editing. 

J.N.K.: conceptualization, funding acquisition, methodology, project administration, supervision, writing-review and editing. 

S.L.B.: conceptualization, funding acquisition, methodology, project administration, supervision, writing-review and editing.

\bibliographystyle{elsarticle-num} 
\bibliography{KneiflEtAl24Multi.bib}

\end{document}

%% file: mh_variables.tex
\usepackage{bm}
\usepackage{amsmath}
\usepackage{xcolor}
\usepackage{xspace}
\usepackage{mathtools}
\usepackage{mathrsfs}

\newcommand{\transpose}{\intercal}
\newcommand{\eye}{\bm{I}}
\newcommand{\Fourier}{\mathscr{F}}

\newcommand{\diag}{\text{diag}}
\newcommand{\ddt}{\frac{\mathrm{d}}{\mathrm{d}t}}

\newcommand{\mean}[1]{\underset{#1}{\text{mean}}\,}

\newcommand{\fourier}[1]{\hat{#1}}
\newcommand{\approximate}[1]{\tilde{#1}}
\newcommand{\reconstruct}[1]{\breve{#1}}
\newcommand{\scale}[1]{\check{#1}}
\newcommand{\Flow}{\bm{F}}
\newcommand{\rhs}{\bm{f}}

\newcommand{\Rdim}{\ensuremath\mathbb{R}}
\newcommand{\binare}{\ensuremath\{0,1\}}

\newcommand{\stateScalar}{x}
\newcommand{\states}{\bm{\stateScalar}}
\newcommand{\stateApprox}{\approximate{\states}}
\newcommand{\stateRec}{\reconstruct{\states}}
\newcommand{\stateSpace}{\mathcal{X}}

\newcommand{\stateDim}{N}
\newcommand{\snaps}{\bm{X}}

\newcommand{\redStateDim}{r}
\newcommand{\redStateSpace}{\mathcal{Z}}
\newcommand{\redState}{\bm{z}}
\newcommand{\redStateApprox}{\tilde{\redState}}

\newcommand{\param}{\mu}
\newcommand{\params}{\bm{\param}}
\newcommand{\paramSpace}{\mathcal{M}}

\newcommand{\paramDim}{\ell}
\newcommand{\timee}{t}
\newcommand{\timeSpace}{\mathcal{T}}
\newcommand{\nTime}{\eta}
\newcommand{\nSims}{n_s}

\newcommand{\nodeIteratorA}{p}
\newcommand{\nodeIteratorB}{q}

\newcommand{\simIterator}{s}

\newcommand{\solutionManifold}{\mathcal{S}}

\newcommand{\disp}{q}
\newcommand{\disps}{\bm{\disp}}

\newcommand{\compTime}{\Delta T}

\newcommand{\surrogate}{\bm{\Sigma}}
\newcommand{\regression}{\bm{\Phi}}
\newcommand{\reduction}{\bm{\Psi}_{\text{enc}}}
\newcommand{\reductionSubscript}[1]{\bm{\Psi}_{\text{enc}, #1}}
\newcommand{\reconstruction}{\bm{\Psi}_{\text{dec}}}
\newcommand{\reconstructionSubscript}[1]{\bm{\Psi}_{\text{dec}, #1}}

\newcommand{\eigenvalue}{\lambda}
\newcommand{\eigenvalueMatrix}{\bm{\Lambda}}
\newcommand{\eigenvector}{\bm{u}}
\newcommand{\eigenvectors}{\bm{U}}

\newcommand{\down}{\text{d}}
\newcommand{\up}{\text{u}}

\newcommand{\nodeComponent}{\nu}
\newcommand{\node}{\bm{\nodeComponent}}
\newcommand{\nodes}{\bm{\mathcal{N}}}
\newcommand{\adjacency}{\bm{A}}
\newcommand{\laplacian}{\bm{L}}
\newcommand{\degreeMatrix}{\bm{D}}
\newcommand{\nodesDown}{\nodes_\down}
\newcommand{\nodesUp}{\nodes_\up}
\newcommand{\downsampling}{\mathcal{\bm{D}}}
\newcommand{\upsampling}{\mathcal{\bm{U}}}
\newcommand{\samplingNetwork}{\bm{\varTheta}}
\newcommand{\quadric}{\bm{Q}}

\newcommand{\nNodes}{n}
\newcommand{\nEdges}{n_{\text{e}}}
\newcommand{\nNodesDown}{n_{\down}}

\newcommand{\channelIterator}{c}
\newcommand{\nChannels}{n_{c}}
\newcommand{\chebIterator}{k}
\newcommand{\chebOrder}{K}
\newcommand{\chebPoly}{T}

\newcommand{\mlInput}{x}

\newcommand{\mlInputs}{\bm{\mlInput}}
\newcommand{\mlInputMatrix}{\bm{X}}

\newcommand{\nlevel}{\ell}

\newcommand{\loss}{\mathcal{L}}
\newcommand{\lossWeights}{\gamma}
\newcommand{\dataSet}{\mathcal{D}}
\newcommand{\weight}{{w}}
\newcommand{\weights}{\bm{\weight}}
\newcommand{\Weights}{\bm{W}}
\newcommand{\layerIterator}{l}

\newcommand{\activationFunction}{\bm{h}}

\newcommand{\graph}{\mathcal{G}}

\newcommand{\edges}{\mathcal{E}}
\newcommand{\filter}{\bm{g}}
\newcommand{\filterMatrix}{\filter_{\weights}}

\newcommand{\error}{\ensuremath e}
\newcommand{\eTwo}{\ensuremath e_{\text{2}}}

\newcommand{\eTwoMean}{\ensuremath \hat{\eTwo}}

\newcommand{\FE}{\textsf{FE}\xspace}

\newcommand{\MOR}{\textsf{MOR}\xspace}
\newcommand{\PCA}{\textsf{PCA}\xspace}
\newcommand{\POD}{\textsf{POD}\xspace}

\newcommand{\MH}{\textsf{MH}\xspace}
\newcommand{\MLP}{\textsf{MLP}\xspace}

\newcommand{\CNNs}{\textsf{CNNs}\xspace}

\newcommand{\GCNNs}{\textsf{GCNNs}\xspace}

\renewcommand{\AE}{\textsf{AE}\xspace}

%% file: fig/figure_config.tex
\usepackage{bm}
\usepackage{amsmath}
\usepackage{xcolor}
\usepackage{pgfplots}
\usetikzlibrary{calc,
        positioning, 
        shapes.geometric, 
        math, 
        arrows.meta, 
        fadings, 
        through, 
        quotes, 
        patterns, 
        pgfplots.groupplots
}

\definecolor{britishracinggreen}{rgb}{0.0, 0.26, 0.15}
\definecolor{midnightblue}{rgb}{0.1, 0.1, 0.44}
\definecolor{darkpastelblue}{rgb}{0.47, 0.62, 0.8}
\definecolor{darkmidnightblue}{rgb}{0.0, 0.2, 0.4}

\definecolor{recColor}{rgb}{0,0.419607843137255,0.643137254901961}
\definecolor{regrColor}{rgb}{0.88,0.48,0.62} 
\colorlet{latentColor}{purple} 
\definecolor{arsenic}{rgb}{0, 0.5, 0.5}

\definecolor{approxColor}{rgb}{1,1,0.0549019607843137}
\definecolor{applegreen}{rgb}{0.55, 0.71, 0.0}
\definecolor{musclecolor}{rgb}{0.8666666666666667, 0.5411764705882353, 0.5411764705882353}

\definecolor{fineColor}{rgb}{0.3,0.3,0.3}
\definecolor{coarseColor}{rgb}{0.0, 0.29, 0.33}
\definecolor{faultyColor}{rgb}{0.86, 0.52, 0.0}
\definecolor{refinedColor}{rgb}{0.8666666666666667, 0.5411764705882353, 0.5411764705882353}

\colorlet{podColor}{midnightblue}
\colorlet{aeColor}{orange}
\colorlet{gcnColor}{magenta}
\colorlet{mhColor}{arsenic}

\definecolor{mhColor2}{RGB}{0, 180, 180} 
\definecolor{mhColor3}{RGB}{0, 230, 230} 

\tikzset{neuron/.style={circle, draw, minimum size=4ex, thick},}
\tikzset{legend/.style={font=\large\bfseries},}
\tikzset{font/.style={\fontsize{5pt}{12}},}
\tikzset{cell/.style={
        rectangle, 
        rounded corners=1mm, 
        minimum height =0.7cm, 
        minimum width=0.7cm, 
        draw,
        thick,
        anchor=center,
},}
\tikzset{function/.style={
    rectangle, 
    rounded corners=1mm, 
    minimum height =1cm, 
    minimum width=1.5cm, 
    draw,
    thick,
    anchor=center,
    },}
\tikzset{mylabel/.style={%
  align=center
},}
\tikzset{Arrow/.style={
        rounded corners=.1cm,
        thick,
        shorten >= 3pt,
        shorten <= 3pt
},}
\tikzset{reduc/.style={
        trapezium,
        rounded corners=0.5mm,
        shape border 
        rotate=270, 
        trapezium stretches=true,
        trapezium angle=45, 
        minimum height=10mm, 
        minimum width=22mm, 
        inner sep=0mm,
        draw=recColor, 
        fill=recColor!40!white,
        thick, 
        anchor=west,}}

\tikzset{rec/.style={
        trapezium,
        rounded corners=0.5mm,
        shape border 
        rotate=90, 
        trapezium stretches=true,
        trapezium angle=45, 
        minimum height=10mm, 
        minimum width=22mm, 
        draw=recColor, 
        fill=recColor!40!white,
        inner sep=0mm,
        thick,  
        anchor=west}}

\tikzset{upsampling/.style={
        rec,
        trapezium angle=82, 
        minimum height=10mm, 
        minimum width=32mm, 
        draw=\samplingColor, 
        fill=\samplingColor!40!white,}}

\tikzset{downsampling/.style={
        reduc,
        trapezium angle=82, 
        minimum height=10mm, 
        minimum width=32mm, 
        draw=\samplingColor, 
        fill=\samplingColor!40!white,}}

\newcommand\connectionsFromAdjacency[3]{
  \begin{pgfonlayer}{bg2}    
    \foreach [count=\r] \row in #3{
      \foreach [count=\c] \cell in \row{
          \ifnum\cell=1%
              \draw[arc/.try=\cell, 
                bend left=00, 
                draw=#2, 
                line width=0.3mm,
                ] (#1-\r) edge (#1-\c);
          \fi
      }
    }
  \end{pgfonlayer}
}

\newcommand\drawNodes[4]{
  \foreach [count=\i] \x/\y/\color/\d in #2  {
      \pgfmathsetmacro\randCol{random(20,70)}
        \ifnum\d=1%
          \node[circle,
                inner sep=#4pt, 
                fill=\color!\randCol!white, 
                draw=\color,
            ] 
            (#1-\i) at (10*\x,10*\y) {};
          \foreach \j in #3 {
            \pgfmathsetmacro\randCol{random(20,70)}
            \node[circle,
                inner sep=#4pt,
                fill=\color!\randCol!white, opacity=.8, 
                draw=\color
                ] 
              (#1h-\i\j) at ($(10*\x,10*\y)+(\j*0.05, -\j*0.05)$) {};
          }
        \fi
  }
}

\pgfdeclarelayer{bg0}    
\pgfdeclarelayer{bg1}    
\pgfdeclarelayer{bg2}    
\pgfsetlayers{bg0, bg1, bg2, main}